\documentclass[10pt,twocolumn,letterpaper]{article}

\usepackage{cvpr}              % To produce the CAMERA-READY version
\usepackage[dvipsnames,table,xcdraw]{xcolor}

\definecolor{cvprblue}{rgb}{0.21,0.49,0.74}
\usepackage[pagebackref,breaklinks,colorlinks,citecolor=cvprblue]{hyperref}
\usepackage{booktabs}
\usepackage{graphicx}
\usepackage{multirow}
\usepackage{algorithm}
\usepackage{algorithmic}
\usepackage[accsupp]{axessibility}

\makeatletter
\def\customsymbol#1{
    \ifcase\number\value{#1}
        \or *
        \or \Letter
    \else\@ctrerr
    \fi
}
\makeatother

\def\paperID{10293} % *** Enter the Paper ID here
\def\confName{CVPR}
\def\confYear{2024}

\title{Orchestrate Latent Expertise: Advancing Online Continual Learning with Multi-Level Supervision and Reverse Self-Distillation}

\author{
    Hongwei Yan$^{1}$\qquad Liyuan Wang$^{2*}$\qquad Kaisheng Ma$^{3*}$\qquad Yi Zhong$^{1}$\thanks{Corresponding authors.} \\
    $^{1}$ School of Life Sciences, IDG/McGovern Institute for Brain Research, Tsinghua University\\
    $^{2}$ Dept. of Comp. Sci. \& Tech., Institute for AI, BNRist Center,\\ Tsinghua-Bosch Joint ML Center, THBI Lab, Tsinghua University\\
    $^{3}$ Institute for Interdisciplinary Information Sciences, Tsinghua University\\
    {\tt\small yanhw22@mails.tsinghua.edu.cn},\quad
    {\tt\small \{wly2023, kaisheng, zhongyithu\}@tsinghua.edu.cn}
}

\begin{document}
\maketitle
\begin{abstract}

\setcounter{footnote}{1}
\renewcommand{\thefootnote}{\fnsymbol{footnote}}

To accommodate real-world dynamics, artificial intelligence systems need to cope with sequentially arriving content in an online manner. Beyond regular Continual Learning (CL) attempting to address catastrophic forgetting with offline training of each task, Online Continual Learning (OCL) is a more challenging yet realistic setting that performs CL in a one-pass data stream. Current OCL methods primarily rely on memory replay of old training samples. However, a notable gap from CL to OCL stems from the additional overfitting-underfitting dilemma associated with the use of rehearsal buffers: the inadequate learning of new training samples (underfitting) and the repeated learning of a few old training samples (overfitting). To this end, we introduce a novel approach, Multi-level Online Sequential Experts (MOSE), which cultivates the model as stacked sub-experts, integrating multi-level supervision and reverse self-distillation. Supervision signals across multiple stages facilitate appropriate convergence of the new task while gathering various strengths from experts by knowledge distillation mitigates the performance decline of old tasks. MOSE demonstrates remarkable efficacy in learning new samples and preserving past knowledge through multi-level experts, thereby significantly advancing OCL performance over state-of-the-art baselines (\eg, up to 7.3\% on Split CIFAR-100 and 6.1\% on Split Tiny-ImageNet). 
\footnote{Our code is available at \href{https://github.com/AnAppleCore/MOSE}{https://github.com/AnAppleCore/MOSE}}

\end{abstract}
\section{Introduction}
\label{sec:intro}

% \begin{figure}[]
%   % \vspace{-0.1cm}
%   \centering
%   \fbox{\rule{0pt}{1.5in} \rule{.9\linewidth}{0pt}}
%   \caption{\textbf{Intuition}. intuition
%   }
%   \label{fig:intuition}
%   % \vspace{-0.5cm}
% \end{figure}

Continual Learning (CL) aims to improve the adaptability of AI systems to the ever-evolving environment of the real world~\cite{Zhou2023CLreview, lange2022CLreview, liyuan2023CLreview, Masana2023CLreview, Ignacio2019CLreview}. In this pursuit, Online Continual Learning (OCL) has emerged as an important paradigm that mirrors realistic scenarios with one-pass data streams and attracted a wide range of interest in related fields~\cite{He2020OCLreview, mai2022OCLreview, Albin2023OCLreview}. However, OCL also faces serious challenges such as efficient online operation, limited input data, and stringent resource constraints, which remain to be solved.

In this work, we first analyze the unique requirements of OCL compared to regular CL:
the training samples for each task are encountered only once, which makes the model susceptible to inadequate learning of each task~\cite{MuFAN}. Saving samples of the current task in the buffer could partially solve it, but overstepping the training of buffered data will trigger a severe forgetting problem, namely, overfitting to the memory buffer for old tasks~\cite{RAR, RR}. This critical issue is attributed to the \textbf{overfitting-underfitting dilemma} across data distributions of new and old tasks, which specifies the distinction between CL and OCL. While recent strides have been made with the help of data augmentation~\cite{IL2A, RAR} and contrastive representation learning~\cite{SCR, PCR, CCPR, cheng2023contrastive}, shortcomings such as the sub-optimality of convergence and the demand for large batch size with higher computational overhead remind us a chasm in performance persists~\cite{Albin2023OCLreview}.

Unlike AI systems, the biological brain has evolved an innate capacity for continual learning in an online manner. In particular, the mammalian visual processing system extracts multi-level features of online inputs and memorizes them for further reuse~\cite{human_viual_cortex, primary_visual_cortex, ventral_visual_pathway, human_ventral_visual_pathway}. Neurons in these areas not only process and transmit signals progressively but also communicate across cortex levels through diverse neural circuits~\cite{cossell2015functional}. The \textit{Shallow Brain} hypothesis~\cite{suzuki2023deep} is proposed to describe the ability of the superficial cortex to work independently and cooperate with deeper ones. Inspired by this, we propose the \textbf{M}ulti-level \textbf{O}nline \textbf{S}equential \textbf{E}xperts (MOSE), a novel approach to drive the leap forward in OCL. 
MOSE consists of two major components: multi-level supervision and reverse self-distillation. The former empowers the model to forge hierarchical features across various scales, cultivating the continual learner as stacked sub-experts excelling at different tasks. Meanwhile, the latter shifts knowledge within the model from shallower experts to the final predictor, gathering the essence of diverse expertise. Combined with both modifications, the model tackles tasks by orchestrating a harmonious symphony of latent network skills, endowed with resistance against distribution shift and task-wise accuracy.

To address the identified particular challenge of OCL, MOSE places the model at an appropriate convergence point to facilitate efficient learning of new tasks, while avoiding performance degradation of old tasks. 
The cooperation of multi-level experts achieves such a flexible balance between overfitting and underfitting. As a result, our MOSE substantially outperforms state-of-the-art baselines.

Our contributions can be summarized as three aspects:
\begin{enumerate}
    \item We present an extensive analysis of the OCL problem and attribute its particular challenge to the overfitting-underfitting dilemma of the observed data distributions. 
    \item We propose an innovative approach with multi-level supervision and reverse self-distillation, to achieve appropriate convergence in an online manner.
    \item Empirical experiments demonstrate the superior performance of MOSE over state-of-the-art baselines.
\end{enumerate}
\section{Related Work}
\label{sec:related}

\textbf{Continual Learning (CL).}
CL has received increasing attention in recent years~\cite{Zhou2023CLreview, liyuan2023CLreview, Masana2023CLreview}, characterized by non-stationary data learning. Conventionally, CL methods are classified into three groups: architecture-, regularization- and replay-based~\cite{ASER, Ignacio2019CLreview, lange2022CLreview}. Architecture-based methods focus on allocating dedicated parameter subspace, including parameter isolation~\cite{hardattn, SupSup}, dynamic architecture~\cite{DGM, hypernetwork, wang2024hierarchical, gao2024enhancing}, and modular network~\cite{coscl,wang2023incorporating}. Regularization-based methods~\cite{Kirkpatrick2017overcomecatastrophic, CAB, wang2021afec, lyu2024overcoming} mitigate catastrophic forgetting by introducing explicit regularization terms to balance new and old tasks. Replay-based methods~\cite{DER, AGEM, GDumb, wang2021ordisco, MRDC} exploit an additional memory buffer to save a subset of old training samples, to recover previous data distributions.
As regular CL methods usually perform multi-epoch training of each task, it remains extremely challenging to deal with the one-pass data stream in OCL~\cite{OCM}.

\noindent
\textbf{Online Continual Learning (OCL).}
In OCL, a model needs to learn a one-pass data stream with shifting distribution~\cite{He2020OCLreview, mai2022OCLreview, Albin2023OCLreview}.
Replay-based methods have been extensively explored in OCL thanks to their efficacy and generality~\cite{Albin2023OCLreview}.
ER~\cite{ER} applies a reservoir sampling strategy~\cite{Reservoir} and randomly updates the memory buffer. MIR~\cite{MIR} selects the most interfered replay samples. SCR~\cite{SCR} takes supervised contrastive loss and nearest-class-mean classifier. ER-AML~\cite{ER-AML} modifies cross-entropy loss to mitigate representation drift. OCM~\cite{OCM} learns holistic features through maximizing mutual information. OnPro~\cite{OnPro} avoids short-cut learning using online prototypes. GSA~\cite{GSA} improves cross-task discrimination using a gradient-based method.
Specifically, we adopt data augmentation for its capacity to expand the data distribution across both stored and incoming batches. The combination of \emph{memory replay} and \emph{data augmentation} currently dominates the OCL literature, yet architecture-based methods remain to be explored in OCL.
Our method integrates their strengths and is compatible with mainstream replay-based methods.

\noindent
\textbf{Knowledge Distillation (KD) in CL.} 
KD~\cite{Hinto2015KD, ba2014kd} usually saves a frozen copy of the old model to ``teach'' the current model.
According to the target space, there usually exist logits-based, feature-based, and relation-based KD methods~\cite{Zhou2023CLreview}. LwF~\cite{LWF} learns outputs from the previous model; DER~\cite{DER} employs logit-based distillation, where previous logits are saved in memory buffer; iCaRL~\cite{iCaRL} preserves learned representations through distillation; CO$^2$L~\cite{CO2L} employs self-supervised distillation to keep robust representations; MuFAN~\cite{MuFAN} introduces structure-wise distillation to build relation between tasks and CCPR~\cite{CCPR} transfers feature correlation between training samples from past model.
In contrast, our method innovates by implementing KD internally within the current model architecture, setting it apart from conventional KD techniques.
\section{Preliminaries}
\label{sec:prelinminary}

% In this section, we first introduce the OCL setup and representative methods and then present an in-depth empirical analysis of its particular challenge.

\subsection{Notation and Setups}
\label{sec:notation}

\textbf{Problem Description.} Let's consider a general setting of OCL, which usually refers to fitting a one-pass data stream with non-stationary distribution~\cite{He2020OCLreview, mai2022OCLreview, Albin2023OCLreview}. 
The training dataset $\mathcal{D}$ is split into several tasks, $\mathcal{D}_{t\leq T} = \{(x_i,y_i)|y_i\in \mathcal{C}_t\}$, and $\bigcup_{t=1}^{T}\mathcal{C}_t= \mathcal{C}$ is the set of labels, where the sets of class labels are disjoint $\mathcal{C}_t \cap \mathcal{C}_{t^\prime} = \emptyset$.
The continual part requires that task training sets arrive sequentially and are not accessible without additional cost, whenever the training phase of the corresponding task is complete. The unavailability of past data causes a biased input distribution and therefore induces catastrophic forgetting~\cite{MCCLOSKEY1989catastrophic, french1999catastrophic, goodfellow2015catastrophic} of old tasks. This challenge is further strengthened for OCL: during the training phase of task $t$, sequential data batches $\mathcal{B}^t = \{(x_i, y_i)\}_{i=1}^B$ of batch size $B$ are sampled from its training set $\mathcal{D}_t$ in a no replacement manner. Models are restricted to process each training sample (input-label pair) precisely once -- \textbf{at most one epoch over each training set}. Thus, the continual learner also struggles to learn each new task.
To tackle this challenge, OCL methods frequently employ memory replay strategy~\cite{OCM, OnPro}.
They apply a memory buffer $\mathcal{M}$ with limited capacity $M$ to store old training samples. At each training iteration, a batch of data $\mathcal{B}^\mathcal{M} = \{(x_i, y_i)\}_{i=1}^{B^\mathcal{M}}$ of batch size $B^\mathcal{M}$ is retrieved from the memory buffer $\mathcal{M}$, and the model is updated according to the loss calculated over $\mathcal{B}^t$ and $\mathcal{B}^\mathcal{M}$ (detailed in Sec.~\ref{sec:replay_approaches}).

\noindent
\textbf{Model Architecture.} Implemented as multiple consecutive modules, deep learning models depend heavily on the power of network depth. Backbones commonly used in computer vision like ResNet~\cite{resnet} and ViT~\cite{VIT} consist of stacked repetitive blocks such as convolution and multi-headed self-attention. A general network $F:\mathbb{R}^{m}\mapsto \mathbb{R}^{|\mathcal{C}|}$ maps an input $x$ of size $m$ to a class probability vector $\hat{y}$ for classification problem, consisting of a feature extractor $f_\theta$ and an output layer $g_{\phi}$ with parameters $\theta$, $\phi$. 
The feature extractor $f_\theta$ includes a series of successively connected blocks and encodes $x$ to a feature vector of dimension $d$:
\begin{small}
\begin{equation}
\begin{split}
    f_\theta(x) & = \left(f_{\theta_n}\circ f_{\theta_{n-1}}\circ \cdots \circ f_{\theta_{1}} \right) (x) \in \mathbb{R}^d, \\
    \theta &=\{\theta_1, \theta_2, \cdots, \theta_n\},
\end{split}    
\label{eq:consective_blocks}
\end{equation}
\end{small}
where each block $f_{\theta_{i}}$ takes the feature map $h_{i-1}$ from its predecessor $f_{\theta_{i-1}}$ and produces a new one $h_{i}=f_{\theta_{i}}(h_{i-1})$. Multi-level feature maps are obtained from blocks with different depths $(h_0 = x$, $h_n = f_\theta(x))$. And a classifier network learns to minimize the classification loss $\mathcal{L}(\hat{y}, y)$ between its output $\hat{y}=F(x;\theta, \phi) = g_{\phi}(h_n) = g_{\phi}\left(f_{\theta} (x)\right)$ and ground-truth $y$. Cross-entropy loss is usually used:
\begin{small}
\begin{equation}
  \mathcal{L}_{ce}(\hat{y}, y) = - \sum_{c\in \mathcal{C}} y^{c} \log\left(
    \frac{\exp(\hat{y}^c)}{\sum_{s\in\mathcal{C}}\exp(\hat{y}^s)}
  \right).
  \label{eq:standard_ce}
\end{equation}
\end{small}
$y^c$ and $\hat{y}^c$ are the probabilities of being classified as $c$.

\subsection{Representative Method}
\label{sec:replay_approaches}

Regarding the memory buffer $\mathcal{M}$, three aspects need to be considered: (1) \textbf{memory buffer size $M$} to determine the number of old training samples that can be stored in the memory buffer; (2) \textbf{memory retrieval strategy}~\cite{MIR, ASER} to gather old training samples from the memory buffer; and (3) \textbf{memory update strategy}~\cite{GSS, food, GMED} to keep an appropriate data structure across tasks and classes. Existing methods~\cite{CBRS, GSS, GMED, CCPR} demonstrate variability reflecting diverse interpretations of the OCL problem and numerous possible solutions. Here we describe two representative baselines:

\noindent
\textbf{Experience Replay (ER).} The most straightforward baseline is the naive ER~\cite{ER}, using reservoir sampling strategy~\cite{Reservoir} to randomly retrieve old samples from the buffer and also update it with random addition and deletion in a balanced manner.
The model $F$ learns incremental batches and old training samples $\mathcal{B} = \mathcal{B}^\mathcal{M} \cup \mathcal{B}^t$ with cross-entropy:
\begin{equation}
    \mathcal{L}_{\text{ER}} = \mathbb{E}_{(x_i,y_i)\in \mathcal{B}} \mathcal{L}_{ce}\left(F(x_i;\theta, \phi), y_i\right).
    \label{eq:er_loss}
\end{equation}

\noindent
\textbf{Supervised Contrastive Replay (SCR).} SCR~\cite{SCR} introduces a supervised contrastive loss that learns more refined feature encoding by drawing the same class closer while pushing different classes away.
A nearest-class-mean (NCM~\cite{NCM}) classifier is used instead of output class probability directly.
During the testing phase, SCR compares the distance between the encoded input feature with class feature means and predicts the closest one. Replacing $g_{\phi}$ with a linear projection layer $p_{\psi}$, SCR computes its loss $\mathcal{L}_{scl}$ between features $q=F(x;\theta, \psi)=p_{\psi}\left(f_{\theta}(x)\right)$ over $\mathcal{B}$:
\begin{small}
\begin{equation}
\begin{split}
    \mathcal{L}_{scl}(q_i, y_i) &= -\mathbb{E}_{p\in\mathcal{P}(i)} \log \left(
        \frac{\exp(q_i \cdot q_p / \tau)}{\sum_{j\neq i}\exp(q_i \cdot q_j / \tau)}
    \right), \\
    \mathcal{L}_{\text{SCR}} &= \mathbb{E}_{(x_i, y_i)\in \mathcal{B}} \mathcal{L}_{scl} (F(x_i;\theta, \psi), y_i),
\end{split}
\label{eq:scr_loss}
\end{equation}
\end{small}
where $\tau$ is a temperature hyperparameter, and $\mathcal{P}(i)$ is the index set of positive samples in $\mathcal{B}$, whose class labels are the same as $x_i$ but excluding $x_i$ itself.

\subsection{Empirical Analysis}
\label{sec:empricial_analysis}

\begin{figure*}[t]
  % \vspace{-0.1cm}
  \centering
  \begin{subfigure}{0.245\linewidth}
    % \fbox{\rule{0pt}{1.5in} \rule{.9\linewidth}{0pt}}
    \includegraphics[width=\linewidth]{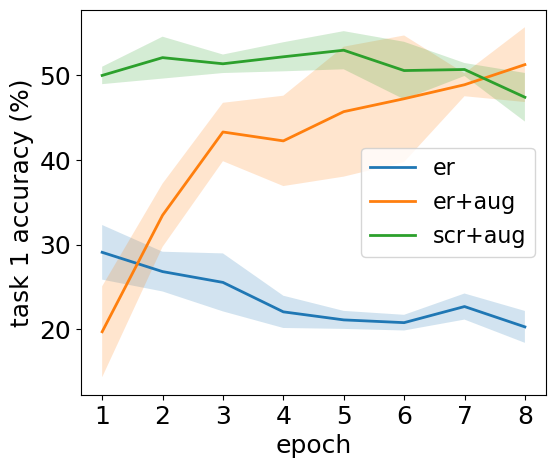}
    \caption{Accuracy of task 1 when $t=2$}
    \label{fig:t1acc@t2}
  \end{subfigure}
  \begin{subfigure}{0.245\linewidth}
    % \fbox{\rule{0pt}{1.5in} \rule{.9\linewidth}{0pt}}
    \includegraphics[width=\linewidth]{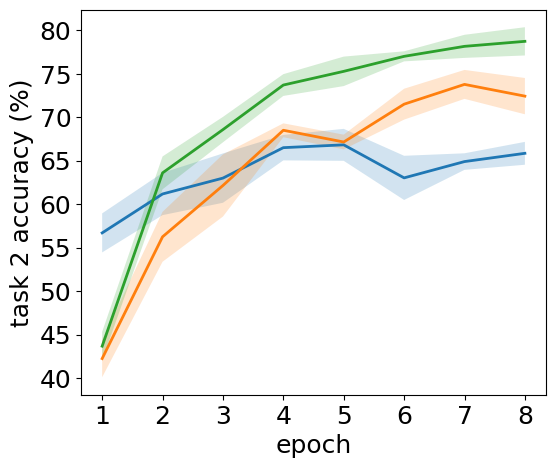}
    \caption{Accuracy of task 2 when $t=2$}
    \label{fig:t2acc@t2}
  \end{subfigure}
  \begin{subfigure}{0.245\linewidth}
    % \fbox{\rule{0pt}{1.5in} \rule{.9\linewidth}{0pt}}
    \includegraphics[width=\linewidth]{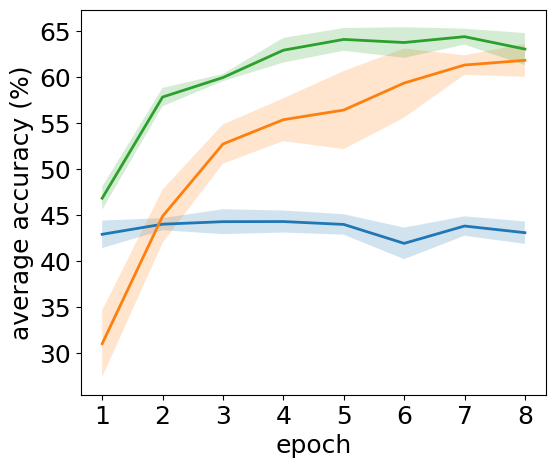}
    \caption{Overall accuracy when $t=2$}
    \label{fig:allacc@t2}
  \end{subfigure}
  \begin{subfigure}{0.245\linewidth}
    % \fbox{\rule{0pt}{1.5in} \rule{.9\linewidth}{0pt}}
    \includegraphics[width=\linewidth]{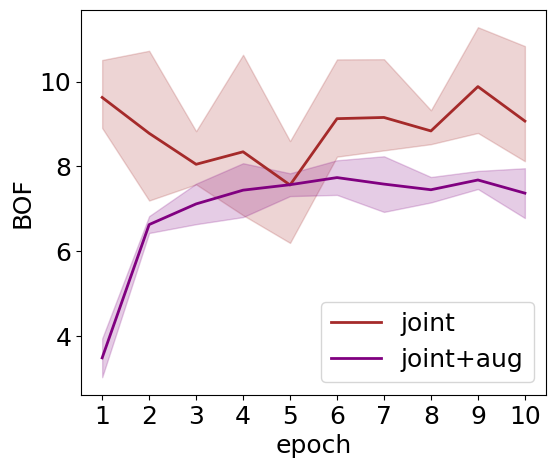}
    \caption{BOF for buffer joint training}
    \label{fig:joint_bof}
  \end{subfigure} \\
  \caption{\textbf{Overfitting-Underfitting Dilemma.} We show the impact of training different epochs to the test accuracy of task~1 and task~2 of Split CIFAR-100~\cite{cifar} dataset, as well as BOF value for joint training on the buffer of the last task. \textbf{(a)} shows the test accuracy of task~1 when the training of task~2 has just finished ($t=2$). Similarly, \textbf{(b)} and \textbf{(c)} show the test accuracy of task~2, and the average performance of the first two tasks when $t=2$. \textbf{(d)} shows the buffer overfitting problem across different epochs. \texttt{aug} here is a combination of 3 different data augmentation used in~\cite{simclr, OCM}. For a fair comparison, we fix $M=1K$, batch size $B=10$, and buffer batch size $B^{\mathcal{M}}=64$.
  }  
  \label{fig:epoch_acc_plot}
    \vspace{-0.3cm}
\end{figure*}

The significant change in data distribution and inadequate training for each current task jointly constitute the central challenge of OCL. Although memory replay is helpful, its assistance is quite limited and costly. For the later issue, previous works~\cite{RAR} point out that, updating multiple times over one incoming batch directly is not an ideal choice because it gradually becomes biased on this small batch.
Data augmentation could enhance the unified data distribution of memory buffer and incoming batches. However, repetitive training still unavoidably brings in much more calculation, and therefore to a certain extent it violates the online requirements of efficiency~\cite{RAR}.
Although contrastive-based representation learning methods exhibit more generalization ability~\cite{CCPR, PCR, DVC, SCR}, the scarcity of available training samples in OCL keeps them from optimal convergence.

To further study the impact of training iterations and data augmentation on the OCL problem, aiming to narrow the performance gap between OCL and offline CL, we conducted a toy experiment to assess the task performance of ER and SCR under different numbers of task epochs and augmentation conditions (see Fig.~\ref{fig:epoch_acc_plot}). We evaluate both task-wise prediction performance and the average accuracy right after completing the second task of Split CIFAR-100~\cite{cifar}, denoted by $t=2$. The results at ${\rm{epoch}}=1$ correspond to OCL, while offline CL has no specified upper limit on epochs. As epochs increase, the training process in OCL is closer to that in offline CL.

Fig.~\ref{fig:epoch_acc_plot} represents the insufficient training of ER over new task: with more epochs of learning, ER performs better with the current task (Fig.~\ref{fig:t2acc@t2}). However, the descending curve in Fig.~\ref{fig:t1acc@t2} and flat curve in Fig.~\ref{fig:allacc@t2} of ER indicate the improvement of task~2 sacrifices the accuracy of task~1. This balance issue is partially resolved by data augmentation (ER+\texttt{aug}): both new and old tasks gain promotion as the epoch number grows. On the other hand, we capture a significant disparity between fewer and more epochs in Fig.~\ref{fig:allacc@t2} (augmentation operation raises the upper limit), showing its capability to ascend to a superior level. Contrastive learning slightly reduced this gap (similar accuracy as ER+\texttt{aug} at epoch=8, but higher starting point at epoch=1) and exhibits better stability (SCR+\texttt{aug} in Fig.~\ref{fig:t1acc@t2},~\ref{fig:allacc@t2}), yet still far from optimal.

Furthermore, it is important to note that, in replay-based methods, buffered data serves as the core mechanism for preserving past knowledge. Those data can be replayed multiple times throughout the training process. (\eg, for an OCL benchmark with 10 tasks, buffered data from task~1 will be retrieved for 9 epochs during subsequent training), so we conduct another toy example to evaluate the generalization gap between the buffer and the test set. Specifically, we train ER only on the buffer of task~10 (contains a balanced number of samples for each class) for different epochs (denoted as joint, see Fig.~\ref{fig:joint_bof}). We design a new metric Buffer Overfitting Factor (\textbf{BOF}) to quantify:
\begin{small}
\begin{equation}
    \text{BOF} = \frac{\text{Buffer accuracy} - \text{Test set accuracy}}{\text{Test set accuracy}}.
    \label{eq:bof}
\end{equation}
\end{small}
The difference in accuracy between the buffer and the test set is normalized by the test set accuracy, providing a measure of overfitting in relative terms for equitable assessment. The lines in Fig.\ref{fig:joint_bof} reveal that even with augmentation, replaying exemplars from a fixed buffer only twice (epoch=2) yields a substantial BOF. As epochs increase, the BOF stabilizes, indicating a pronounced overfitting phenomenon, similar to the non-augmented case.

% \textbf{Remarks.} At this initial stage, we tentatively summarize previous research consensus~\cite{RAR,RR,SCR,Korycki21replay} and our observations from the above experiment analysis: 
% \begin{enumerate}
%     \item Multiple iterations with ordinary replay-based methods is not a good solution to the underfitting problem of current task, since it triggers drastic forgetting of old tasks.
%     \item Data augmentation provides a better choice with consistent behaviors among tasks by broadening the distribution of both buffered and new instances, but exacerbates the difficulty of convergence over incremental tasks.
%     \item Contrastive based replay methods boost the learning along with data augmentation, while still leaving enough room for further improvement.
%     \item Memory replay helps model to review samples in both new and old task so it alleviates the issue of underfitting but may lead to buffer overfitting of old tasks.
% \end{enumerate}

\noindent
\textbf{Overfitting-Underfitting Dilemma.} Based on the above analysis, we pinpoint the inherent challenge of OCL: OCL methods must navigate the underfitting of the current task while simultaneously avoiding overfitting to the buffered data of previous tasks. This dual challenge inspires us to introduce a novel approach, as outlined below.

\section{Method}
\label{sec:method}

\begin{figure*}[t]
  \centering
    % \vspace{-0.1cm}
  \resizebox{0.88\textwidth}{!}{
    \includegraphics[width=\linewidth]{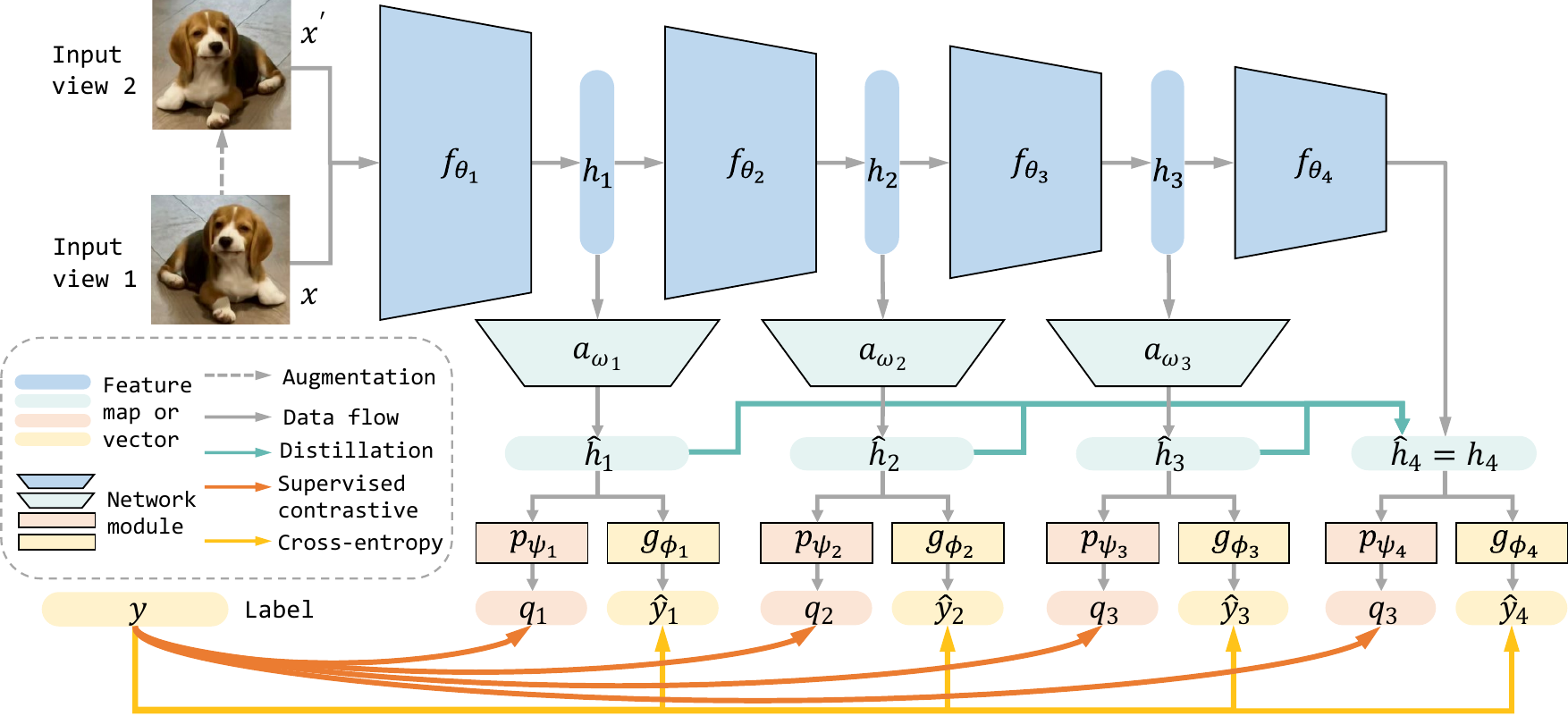}
    }
  \caption{\textbf{Illustration of the proposed MOSE.} For each training sample $(x,y)$, the input $x$ is augmented to another view $x^{\prime}$ and concatenate together for network training. MOSE includes multiple supervision signals (cross-entropy and supervised contrastive loss) injected at different network layers and extra reverse self-distillation from the shallower layers to the deepest to integrate the knowledge of experts.}
  \label{fig:mose}
  \vspace{-0.2cm}
\end{figure*}

Inspired by the hierarchical advantages of the mammalian visual processing system~\cite{human_viual_cortex, human_ventral_visual_pathway, mouse_visual_cortex, ventral_visual_pathway, primary_visual_cortex, ANN}, we propose to inject multi-level supervision in our continual model and aggregate the skills from different experts within it (see Fig.~\ref{fig:mose}), which includes two synergistic components described in Sec.~\ref{sec:multi_level_sup} and Sec.~\ref{sec:reverse_sd}, respectively. Then we summarize the overall framework in Sec.~\ref{sec:overall_mose}.

\subsection{Multi-Level Supervision}
\label{sec:multi_level_sup}

As formulated in Sec.~\ref{sec:notation}, we consider deep neural networks as a composition of sequentially connected building blocks: $f_{\theta_n}\circ f_{\theta_{n-1}}\circ \cdots \circ f_{\theta_{1}}$. The input data is encoded and propagated through the network layer by layer, producing hierarchical feature maps for each training sample. Previous research~\cite{MuFAN} suggested that multi-scale feature maps benefit vision tasks by providing multi-level information ranging from pattern-wise knowledge to highly semantic understanding. This further motivates us to aggregate latent expertise from shallower layers to deeper ones.

First, the whole network can be split into $n$ blocks according to the backbone architecture. Corresponding feature maps $\{h_{i}\in \mathbb{R}^{d_i}\}_{i\leq n}$ are obtained through forward data flow $h_i = f_{\theta_i}(h_{i-1})$, yet they lie in embedding spaces with different dimension $\{d_i\}_{i\leq n}$, ($d_0=m$ is the input size, and $d_n=d$ is the size of final feature to be fed into output head $g_\phi$ or $p_{\psi}$). To unify the feature size for further calculation and comparison, we introduce a dimension alignment module $a_{\omega_i}:\mathbb{R}^{d_i}\mapsto \mathbb{R}^{d}$ with parameter $\omega_i$ to project each feature map to the space with same dimension as the last feature map~\cite{SKD}, namely, $\hat{h}_i = a_{\omega_i}\left(h_i\right) \in \mathbb{R}^{d}.$ These alignment modules serve a dual purpose: they support a general platform for vector operations within hidden space and extract meaningful, diverse information from feature maps that are comparatively sparse and less plausible.

\noindent
\textbf{Latent Sequential Experts.}
Based on the above dimension alignment of multi-scale features, it becomes possible to train each block of the network as a fully functional continual learner. To do so, we add output heads after each block to transform the projected feature $\hat{h}_i$ into an output vector for supervised loss computation. The choice of output head depends on the supervision loss we use, and this multi-level framework is generally compatible with most replay-based methods. Shown in Fig.~\ref{fig:mose}, we add two types of output heads after all alignment module $a_{\omega_i}$: $p_{\psi_i}$ and $g_{\phi_i}$ for supervised contrastive representation learning and cross-entropy-based classification learning (see Sec.~\ref{sec:replay_approaches}), respectively. We then mark them as latent experts $E_i$, consist of sequential blocks $\{f_{\theta_j}\}_{j\leq i}$ up to block $i$ and corresponding feature alignment module $a_{\omega_i}$ and output heads $p_{\psi_i}$ and $g_{\phi_i}$. Naturally, each expert is trained with compounded supervision loss $\mathcal{L}_{E_i}$:
\begin{equation}
\begin{split}
    \mathcal{L}_{E_i} (x, y) & = \mathcal{L}_{ce}\left(\hat{y}_i, y\right) + \mathcal{L}_{scl}\left(q_i, y\right ), \\
    \text{where, } \hat{y}_i & = E_i(x;\theta_{1:i}, \phi) = g_{\phi_i}\left(a_{\omega_i}(f_{\theta_{1:i}}(x))\right), \\
                   q_i & = E_i(x;\theta_{1:i}, \psi) = p_{\psi_i}\left(a_{\omega_i}(f_{\theta_{1:i}}(x))\right).
\end{split}
\label{eq:sub_exp_loss}
\end{equation}
Therefore, \textbf{M}ulti-\textbf{L}evel \textbf{S}upervision (\textbf{MLS}) signal $\mathcal{L}_{\text{MLS}}$ is injected into each block of the network by summing up expert-wise losses and this framework fits perfectly with most replay-based methods:
\begin{equation}
\mathcal{L}_{\text{MLS}} = \mathbb{E}_{(x_i, y_i)\in \mathcal{B}}\sum_{j=i}^n \mathcal{L}_{E_{j}}(x_i, y_i).
\label{eq:mls_loss}
\end{equation}

\noindent
\textbf{Separate $\mathcal{L}_{ce}$ for New Task.}
Recent research indicates that the gradient unbalance observed in output logits of models trained with cross-entropy will lead to bad decision boundaries between old and new tasks~\cite{SSIL, GSA, ER-AML}. From this perspective, we revise the cross-entropy loss $\mathcal{L}_{ce}$ used in expert's loss Eq.~\ref{eq:sub_exp_loss} as $\mathcal{L}_{ce} = \mathcal{L}_{ce, \text{new}}$ + $\mathcal{L}_{ce, \text{buf}}$ to avoid  severe gradient bias caused by repeatedly calculating $\mathcal{L}_{ce}$~\cite{SSIL}. specifically, $\mathcal{L}_{ce, \text{buf}}$ is normally calculated over buffer batch $\mathcal{B}^{\mathcal{M}}$, based on the assumption that memory buffer $\mathcal{M}$ exhibits a better balance across tasks and classes. For incoming batch $\mathcal{B}^t$, loss computed over output logits that represent classes contained in current task $\mathcal{C}^t$:
\begin{equation}
    \mathcal{L}_{ce, \text{new}} (\hat{y}, y) = \sum_{c\in \mathcal{C}^t} y^{c} \log\left(
    \frac{\exp(\hat{y}^c)}{\sum_{s\in\mathcal{C}^t}\exp(\hat{y}^s)}
  \right).
    \label{eq:new_ce_loss}
\end{equation}

\subsection{Reverse Self-Distillation}
\label{sec:reverse_sd}

\begin{table*}[]
\renewcommand\arraystretch{0.75}
  % \vspace{-0.1cm}
\centering
\resizebox{0.9\textwidth}{!}{%
\begin{tabular}{@{}lllllllllllll@{}}
\toprule[2pt]
\multicolumn{2}{c}{\textbf{Experts}} &
  \textbf{task 0} &
  \textbf{task 1} &
  \textbf{task 2} &
  \textbf{task 3} &
  \textbf{task 4} &
  \textbf{task 5} &
  \textbf{task 6} &
  \textbf{task 7} &
  \textbf{task 8} &
  \textbf{task 9} &
  \textbf{Average} \\ \midrule
\multicolumn{1}{l|}{} &
  \multicolumn{1}{l|}{$E_1$} &
  \textbf{50.54} &
  42.85 &
  51.54 &
  41.55 &
  46.36 &
  50.33 &
  48.43 &
  45.51 &
  50.70 &
  \multicolumn{1}{l|}{50.16} &
  47.80 \\
\multicolumn{1}{l|}{} &
  \multicolumn{1}{l|}{$E_2$} &
  47.06 &
  \textbf{45.95} &
  54.47 &
  45.27 &
  50.28 &
  53.87 &
  51.89 &
  50.59 &
  54.70 &
  \multicolumn{1}{l|}{54.41} &
  50.85 \\
\multicolumn{1}{l|}{} &
  \multicolumn{1}{l|}{$E_3$} &
  50.17 &
  45.87 &
  \textbf{55.42} &
  \textbf{45.55} &
  \textbf{50.89} &
  \textbf{55.23} &
  \textbf{53.54} &
  \textbf{53.97} &
  \textbf{59.69} &
  \multicolumn{1}{l|}{61.73} &
  \textbf{53.21} \\
\multicolumn{1}{l|}{} &
  \multicolumn{1}{l|}{\cellcolor[HTML]{EFEFEF}$E_4=F$} &
  \cellcolor[HTML]{EFEFEF}49.70 &
  \cellcolor[HTML]{EFEFEF}44.43 &
  \cellcolor[HTML]{EFEFEF}54.08 &
  \cellcolor[HTML]{EFEFEF}43.95 &
  \cellcolor[HTML]{EFEFEF}49.71 &
  \cellcolor[HTML]{EFEFEF}53.72 &
  \cellcolor[HTML]{EFEFEF}52.78 &
  \cellcolor[HTML]{EFEFEF}53.41 &
  \cellcolor[HTML]{EFEFEF}59.41 &
  \multicolumn{1}{l|}{\cellcolor[HTML]{EFEFEF}\textbf{65.15}} &
  \cellcolor[HTML]{EFEFEF}52.63 \\ \cmidrule(l){2-13} 
\multicolumn{1}{l|}{} &
  \multicolumn{1}{l|}{MAX} &
  50.54 &
  45.95 &
  55.42 &
  45.55 &
  50.89 &
  55.23 &
  53.54 &
  53.97 &
  59.69 &
  \multicolumn{1}{l|}{65.15} &
  53.59 \\
\multicolumn{1}{l|}{\multirow{-6}{*}{w/o RSD}} &
  \multicolumn{1}{l|}{MOE} &
  52.99 &
  48.47 &
  58.28 &
  47.63 &
  53.59 &
  57.33 &
  55.76 &
  56.13 &
  62.78 &
  \multicolumn{1}{l|}{64.23} &
  55.72 \\ \midrule \midrule
\multicolumn{1}{l|}{} &
  \multicolumn{1}{l|}{$E_1$} &
  46.79 &
  43.33 &
  50.74 &
  41.19 &
  46.11 &
  49.80 &
  48.20 &
  45.27 &
  50.02 &
  \multicolumn{1}{l|}{49.66} &
  47.11 \\
\multicolumn{1}{l|}{} &
  \multicolumn{1}{l|}{$E_2$} &
  50.83 &
  46.67 &
  54.11 &
  44.55 &
  49.45 &
  53.85 &
  52.29 &
  49.50 &
  54.75 &
  \multicolumn{1}{l|}{54.21} &
  51.02 \\
\multicolumn{1}{l|}{} &
  \multicolumn{1}{l|}{$E_3$} &
  51.67 &
  47.10 &
  55.39 &
  45.93 &
  51.08 &
  54.79 &
  54.51 &
  53.36 &
  59.71 &
  \multicolumn{1}{l|}{61.13} &
  53.47 \\
\multicolumn{1}{l|}{} &
  \multicolumn{1}{l|}{\cellcolor[HTML]{EFEFEF}$E_4=F$} &
  \cellcolor[HTML]{EFEFEF}\textbf{52.31} &
  \cellcolor[HTML]{EFEFEF}\textbf{48.87} &
  \cellcolor[HTML]{EFEFEF}\textbf{56.75} &
  \cellcolor[HTML]{EFEFEF}\textbf{47.28} &
  \cellcolor[HTML]{EFEFEF}\textbf{51.79} &
  \cellcolor[HTML]{EFEFEF}\textbf{55.72} &
  \cellcolor[HTML]{EFEFEF}\textbf{54.93} &
  \cellcolor[HTML]{EFEFEF}\textbf{54.38} &
  \cellcolor[HTML]{EFEFEF}\textbf{61.45} &
  \multicolumn{1}{l|}{\cellcolor[HTML]{EFEFEF}\textbf{62.26}} &
  \cellcolor[HTML]{EFEFEF}\textbf{54.57} \\ \cmidrule(l){2-13} 
\multicolumn{1}{l|}{} &
  \multicolumn{1}{l|}{MAX} &
  52.31 &
  48.87 &
  56.75 &
  47.28 &
  51.79 &
  55.72 &
  54.93 &
  54.38 &
  61.45 &
  \multicolumn{1}{l|}{62.26} &
  54.57 \\
\multicolumn{1}{l|}{\multirow{-6}{*}{w/ RSD}} &
  \multicolumn{1}{l|}{MOE} &
  53.96 &
  50.29 &
  57.95 &
  48.29 &
  53.31 &
  57.11 &
  55.75 &
  55.23 &
  61.90 &
  \multicolumn{1}{l|}{62.15} &
  55.59 \\ \bottomrule[1.5pt]
\end{tabular}%
  \vspace{-0.2cm}
}
\caption{\textbf{Task-wise End Accuracy (higher is better)} assessed over Split CIFAR-100~\cite{cifar} dataset with memory size $M=5000$. We record the mean result (\%) from 15 runs, and highlight the highest accuracy for each task among 4 sub-experts with \textbf{bold} font. Here $E_4=F$ represents the entire network. \textit{MAX} calculates the ideal situation where we choose the expert with maximal precision for each task, and \textit{MOE} stands for the accuracy of averaged output logits across all experts. We choose NCM classifier~\cite{NCM} for all experts.} 
\label{tab:moe_acc}
\end{table*}

Mixture-of-expert is an effective strategy in machine learning, which cooperates diversified skills of different experts to achieve better predictions~\cite{hinton1993moe, expertgate, coscl}.
A consistent phenomenon is observed when we test the task accuracy for each expert $E_i$. In Tab.~\ref{tab:moe_acc}, the task-wise accuracy of experts, when equipped with MLS, is recorded. Under an ideal situation, we assume that the network can choose the best expert for each task, and the maximum value denoted as \textit{MAX} stands for the upper limit of output from a single expert. We also calculate the accuracy of the average output logits from all experts as a trivial integration process termed as \textit{MOE}.

Based on the result of the first 6 rows of Tab.~\ref{tab:moe_acc}, we conclude experts show differences across tasks, and utilizing multiple experts' outputs surpasses the accuracy of any single expert even if we select the most appropriate one for each task. This further proves the importance of knowledge aggregation across sub-learners. But MOE comes with additional computation and storage costs, and it is not optimal when we have introduced auxiliary modules like feature alignment modules $a_{\omega}$ and extra output heads $p_\psi$, $g_\phi$. Therefore, we intend to transfer skills from different experts to the last one $E_4$, which is indeed the whole network $F$ itself, to avoid redundant overhead at the testing phase.

For this purpose, we implement a novel distillation process called \textbf{R}everse \textbf{S}elf-\textbf{D}istillation (\textbf{RSD}). Different from the traditional self-distillation~\cite{SKD}, our RSD takes latent sequential experts as teachers and treats the largest expert $F$ as the student. It calculates the L2 distance between normalized feature $\hat{h}_i$ (detached to stop gradient back-propagation) of each expert $E_{i<n}$ and the final feature $h_n$ of $E_n$ (Eq.~\ref{eq:rsd_loss}).

The introduction of RSD brings a positive effect on OCL, as shown in Tab.~\ref{tab:moe_acc}.
First, it successfully helps the last expert become the strongest one among all experts without weakening any of them as well as the MOE version. Second, the last expert makes progress on tasks where its initial performance is lacking, contributing positively to OCL challenges. Additional insights on RSD with alternate experts serving as students are detailed in Appendix~\ref{sec:rsd_variations}.
\begin{equation}
\begin{split}
\mathcal{L}_{\text{RSD}} & = \mathbb{E}_{(x_i, y_i)\in \mathcal{B}} \sum_{i=1}^{n-1} \left\|\hat{h}_i^{\prime}-h_n^{\prime}\right\|_2, \\
h^{\prime} & = \text{normalize}(h) = h / \|h\|_2 .
\end{split}
\label{eq:rsd_loss}
\end{equation}

\subsection{Overall Framework of MOSE}
\label{sec:overall_mose}

Put MLS and RSD together, we have established the overall training paradigm for MOSE (see Algo.~\ref{alg:mose} in Appendix~\ref{sec:ipl_details}):
\begin{equation}
\mathcal{L}_{\text{MOSE}} =  \mathcal{L}_{\text{MLS}} + \mathcal{L}_{\text{RSD}}.
\label{eq:mose_loss}
\end{equation}
MOSE is highly generalized and possible to be assembled with most replay-based methods (replace $\mathcal{L}_{ce}$ and $\mathcal{L}_{scl}$ with others) due to its innovative focus on network structure. We then empirically demonstrate the superiority of MOSE and the effectiveness of its building blocks in addressing OCL.

\begin{table*}
\centering
\resizebox{\textwidth}{!}{%
\begin{tabular}{@{}r|cccccc|cccccc@{}}
\toprule[2pt]
\textbf{Method} &
  \multicolumn{6}{c|}{\textbf{Split CIFAR-100 (10 tasks)}} &
  \multicolumn{6}{c}{\textbf{Split Tiny-ImageNet (100 tasks)}} \\ \midrule
Memory Size &
  \multicolumn{2}{c|}{$M=1k$} &
  \multicolumn{2}{c|}{$M=2k$} &
  \multicolumn{2}{c|}{$M=5k$} &
  \multicolumn{2}{c|}{$M=2k$} &
  \multicolumn{2}{c|}{$M=4k$} &
  \multicolumn{2}{c}{$M=10k$} \\ \midrule
Metric &
  ACC(\%) $\uparrow$ &
  \multicolumn{1}{c|}{{\color[HTML]{656565} AF(\%) $\downarrow$}} &
  ACC(\%) $\uparrow$ &
  \multicolumn{1}{c|}{{\color[HTML]{656565} AF(\%) $\downarrow$}} &
  ACC(\%) $\uparrow$ &
  {\color[HTML]{656565} AF(\%) $\downarrow$} &
  ACC(\%) $\uparrow$ &
  \multicolumn{1}{c|}{{\color[HTML]{656565} AF(\%) $\downarrow$}} &
  ACC(\%) $\uparrow$ &
  \multicolumn{1}{c|}{{\color[HTML]{656565} AF(\%) $\downarrow$}} &
  ACC(\%) $\uparrow$ &
  {\color[HTML]{656565} AF(\%) $\downarrow$} \\ \midrule
AGEM(2019) &
  5.8±0.2 &
  \multicolumn{1}{c|}{{\color[HTML]{656565} 77.6±2.0}} &
  5.9±0.3 &
  \multicolumn{1}{c|}{{\color[HTML]{656565} 76.9±1.5}} &
  6.1±0.4 &
  {\color[HTML]{656565} 78.3±1.2} &
  0.9±0.1 &
  \multicolumn{1}{c|}{{\color[HTML]{656565} 73.9±0.2}} &
  2.0±0.5 &
  \multicolumn{1}{c|}{{\color[HTML]{656565} 77.9±0.2}} &
  3.9±0.2 &
  {\color[HTML]{656565} 74.1±0.3} \\
ER(2019) &
  15.7±0.3 &
  \multicolumn{1}{c|}{{\color[HTML]{656565} 66.1±1.3}} &
  21.3±0.5 &
  \multicolumn{1}{c|}{{\color[HTML]{656565} 59.3±0.9}} &
  28.8±0.8 &
  {\color[HTML]{656565} 60.0±1.6} &
  4.7±0.5 &
  \multicolumn{1}{c|}{{\color[HTML]{656565} 68.2±2.8}} &
  10.1±0.7 &
  \multicolumn{1}{c|}{{\color[HTML]{656565} 66.2±0.8}} &
  11.7±0.2 &
  {\color[HTML]{656565} 67.2±0.2} \\
MIR(2019) &
  16.0±0.4 &
  \multicolumn{1}{c|}{{\color[HTML]{656565} 24.5±0.3}} &
  19.0±0.1 &
  \multicolumn{1}{c|}{{\color[HTML]{656565} 21.4±0.3}} &
  24.1±0.2 &
  {\color[HTML]{656565} 21.0±0.1} &
  6.1±0.5 &
  \multicolumn{1}{c|}{{\color[HTML]{656565} 61.1±3.2}} &
  11.7±0.2 &
  \multicolumn{1}{c|}{{\color[HTML]{656565} 60.4±0.5}} &
  13.5±0.2 &
  {\color[HTML]{656565} 59.5±0.3} \\
GSS(2019) &
  11.1±0.2 &
  \multicolumn{1}{c|}{{\color[HTML]{656565} 73.4±4.2}} &
  13.3±0.5 &
  \multicolumn{1}{c|}{{\color[HTML]{656565} 69.3±3.1}} &
  17.4±0.1 &
  {\color[HTML]{656565} 70.9±2.9} &
  3.3±0.5 &
  \multicolumn{1}{c|}{{\color[HTML]{656565} 72.8±1.2}} &
  10.0±0.2 &
  \multicolumn{1}{c|}{{\color[HTML]{656565} 72.6±0.4}} &
  10.5±0.2 &
  {\color[HTML]{656565} 71.5±0.2} \\
ASER(2021) &
  16.4±0.3 &
  \multicolumn{1}{c|}{{\color[HTML]{656565} 25.0±0.2}} &
  12.2±1.9 &
  \multicolumn{1}{c|}{{\color[HTML]{656565} 12.2±1.9}} &
  27.1±0.3 &
  {\color[HTML]{656565} 13.2±0.1} &
  5.3±0.3 &
  \multicolumn{1}{c|}{{\color[HTML]{656565} 65.7±0.7}} &
  8.2±0.2 &
  \multicolumn{1}{c|}{{\color[HTML]{656565} 64.2±0.2}} &
  10.3±0.4 &
  {\color[HTML]{656565} 62.2±0.1} \\
ER-AML(2022) &
  16.1±0.4 &
  \multicolumn{1}{c|}{{\color[HTML]{656565} 51.5±0.8}} &
  17.6±0.5 &
  \multicolumn{1}{c|}{{\color[HTML]{656565} 49.2±0.5}} &
  22.6±0.1 &
  {\color[HTML]{656565} 38.7±0.6} &
  5.4±0.2 &
  \multicolumn{1}{c|}{{\color[HTML]{656565} 47.4±0.5}} &
  7.1±0.5 &
  \multicolumn{1}{c|}{{\color[HTML]{656565} 43.2±0.3}} &
  10.1±0.4 &
  {\color[HTML]{656565} 41.0±0.5} \\
GDumb(2020) &
  17.1±0.4 &
  \multicolumn{1}{c|}{{\color[HTML]{656565} 16.7±0.5}} &
  25.1±0.2 &
  \multicolumn{1}{c|}{{\color[HTML]{656565} 17.6±0.2}} &
  38.6±0.5 &
  {\color[HTML]{656565} 16.8±0.4} &
  12.6±0.1 &
  \multicolumn{1}{c|}{{\color[HTML]{656565} \textbf{15.9±0.5}}} &
  12.7±0.3 &
  \multicolumn{1}{c|}{{\color[HTML]{656565} \textbf{14.6±0.3}}} &
  15.7±0.2 &
  {\color[HTML]{656565} \textbf{11.7±0.2}} \\
SCR(2021) &
  27.3±0.4 &
  \multicolumn{1}{c|}{{\color[HTML]{656565} 17.5±0.2}} &
  30.8±0.5 &
  \multicolumn{1}{c|}{{\color[HTML]{656565} 11.6±0.5}} &
  36.5±0.3 &
  {\color[HTML]{656565} 5.6±0.4} &
  12.6±1.1 &
  \multicolumn{1}{c|}{{\color[HTML]{656565} 19.4±0.3}} &
  18.2±0.1 &
  \multicolumn{1}{c|}{{\color[HTML]{656565} \textit{15.4±0.3}}} &
  21.1±1.1 &
  {\color[HTML]{656565} 14.9±0.7} \\
OCM(2022) &
  28.1±0.3 &
  \multicolumn{1}{c|}{{\color[HTML]{656565} \textit{12.2±0.3}}} &
  35.0±0.4 &
  \multicolumn{1}{c|}{{\color[HTML]{656565} \textit{8.5±0.3}}} &
  42.4±0.5 &
  {\color[HTML]{656565} \textbf{4.5±0.3}} &
  15.7±0.2 &
  \multicolumn{1}{c|}{{\color[HTML]{656565} 23.5±1.9}} &
  21.2±0.4 &
  \multicolumn{1}{c|}{{\color[HTML]{656565} 21.0±0.3}} &
  27.0±0.3 &
  {\color[HTML]{656565} 18.6±0.5} \\
OnPro(2023) &
  30.0±0.4 &
  \multicolumn{1}{c|}{{\color[HTML]{656565} \textbf{10.4±0.5}}} &
  35.9±0.6 &
  \multicolumn{1}{c|}{{\color[HTML]{656565} \textbf{6.1±0.6}}} &
  41.3±0.5  &
  {\color[HTML]{656565} \textit{5.3±0.6}} &
  16.9±0.4 &
  \multicolumn{1}{c|}{{\color[HTML]{656565} \textit{17.4±0.4}}} &
  22.1±0.4 &
  \multicolumn{1}{c|}{{\color[HTML]{656565} 16.8±0.4}} &
 29.8±0.5 &
  {\color[HTML]{656565} 14.6±0.3} \\
GSA(2023) &
  31.4±0.2 &
  \multicolumn{1}{c|}{{\color[HTML]{656565} 33.2±0.6}} &
  39.7±0.6 &
  \multicolumn{1}{c|}{{\color[HTML]{656565} 22.8±0.4}} &
  49.7±0.2 &
  {\color[HTML]{656565} 8.7±0.3} &
  18.4±0.4 &
  \multicolumn{1}{c|}{{\color[HTML]{656565} 35.5±0.3}} &
  26.0±0.2 &
  \multicolumn{1}{c|}{{\color[HTML]{656565} 25.8±0.4}} &
  33.2±0.4 &
  {\color[HTML]{656565} 16.9±0.6} \\ \midrule \midrule
DER++(2020) &
  15.3±0.2 &
  \multicolumn{1}{c|}{{\color[HTML]{656565} 43.4±0.2}} &
  19.7±1.5 &
  \multicolumn{1}{c|}{{\color[HTML]{656565} 44.0±1.9}} &
  27.0±0.7 &
  {\color[HTML]{656565} 25.8±3.5} &
  4.5±0.3 &
  \multicolumn{1}{c|}{{\color[HTML]{656565} 67.2±1.7}} &
  10.1±0.3 &
  \multicolumn{1}{c|}{{\color[HTML]{656565} 63.6±0.3}} &
  17.6±0.5 &
  {\color[HTML]{656565} 55.2±0.7} \\
IL2A(2021) &
  18.2±1.2 &
  \multicolumn{1}{c|}{{\color[HTML]{656565} 24.6±0.6}} &
  19.7±0.5 &
  \multicolumn{1}{c|}{{\color[HTML]{656565} 12.5±0.7}} &
  22.4±0.2 &
  {\color[HTML]{656565} 20.0±0.5} &
  5.5±0.7 &
  \multicolumn{1}{c|}{{\color[HTML]{656565} 65.5±0.7}} &
  8.1±1.2 &
  \multicolumn{1}{c|}{{\color[HTML]{656565} 60.1±0.5}} &
  11.6±0.4 &
  {\color[HTML]{656565} 57.6±1.1} \\
Co$^2$L(2021) &
  17.1±0.4 &
  \multicolumn{1}{c|}{{\color[HTML]{656565} 16.9±0.4}} &
  24.2±0.2 &
  \multicolumn{1}{c|}{{\color[HTML]{656565} 16.6±0.6}} &
  32.2±0.5 &
  {\color[HTML]{656565} 9.9±0.7} &
  10.1±0.2 &
  \multicolumn{1}{c|}{{\color[HTML]{656565} 60.5±0.5}} &
  15.8±0.4 &
  \multicolumn{1}{c|}{{\color[HTML]{656565} 52.5±0.9}} &
  22.5±1.2 &
  {\color[HTML]{656565} 42.5±0.8} \\
LUCIR(2019) &
  8.6±1.3 &
  \multicolumn{1}{c|}{{\color[HTML]{656565} 60.0±0.1}} &
  19.5±0.7 &
  \multicolumn{1}{c|}{{\color[HTML]{656565} 47.5±0.9}} &
  16.9±0.5 &
  {\color[HTML]{656565} 44.3±0.7} &
  7.6±0.5 &
  \multicolumn{1}{c|}{{\color[HTML]{656565} 46.4±0.7}} &
  9.6±0.7 &
  \multicolumn{1}{c|}{{\color[HTML]{656565} 42.2±0.9}} &
  12.5±0.7 &
  {\color[HTML]{656565} 37.6±0.7} \\
CCIL(2021) &
  18.5±0.3 &
  \multicolumn{1}{c|}{{\color[HTML]{656565} 16.7±0.5}} &
  19.1±0.4 &
  \multicolumn{1}{c|}{{\color[HTML]{656565} 16.1±0.3}} &
  20.5±0.3 &
  {\color[HTML]{656565} 17.5±0.2} &
  5.6±0.9 &
  \multicolumn{1}{c|}{{\color[HTML]{656565} 59.4±0.3}} &
  7.0±0.5 &
  \multicolumn{1}{c|}{{\color[HTML]{656565} 56.2±1.3}} &
  15.2±0.5 &
  {\color[HTML]{656565} 48.9±0.6} \\
BiC(2019) &
  21.2±0.3 &
  \multicolumn{1}{c|}{{\color[HTML]{656565} 40.2±0.4}} &
  36.1±1.3 &
  \multicolumn{1}{c|}{{\color[HTML]{656565} 30.9±0.7}} &
  42.5±1.2 &
  {\color[HTML]{656565} 18.7±0.5} &
  10.2±0.9 &
  \multicolumn{1}{c|}{{\color[HTML]{656565} 43.5±0.5}} &
  18.9±0.3 &
  \multicolumn{1}{c|}{{\color[HTML]{656565} 32.9±0.5}} &
  25.2±0.6 &
  {\color[HTML]{656565} 24.9±0.4} \\
SSIL(2021) &
  26.0±0.1 &
  \multicolumn{1}{c|}{{\color[HTML]{656565} 40.1±0.5}} &
  33.1±0.5 &
  \multicolumn{1}{c|}{{\color[HTML]{656565} 33.9±1.2}} &
  39.5±0.4 &
  {\color[HTML]{656565} 21.7±0.8} &
  9.6±0.7 &
  \multicolumn{1}{c|}{{\color[HTML]{656565} 44.4±0.7}} &
  15.2±1.5 &
  \multicolumn{1}{c|}{{\color[HTML]{656565} 36.6±0.7}} &
  21.1±0.1 &
  {\color[HTML]{656565} 29.0±0.7} \\ \midrule \midrule
\rowcolor[HTML]{EFEFEF} 
{\color[HTML]{000000} MOSE} &
  {\color[HTML]{000000} \textit{35.1±0.4}} &
  \multicolumn{1}{c|}{\cellcolor[HTML]{EFEFEF}{\color[HTML]{656565} 36.9±0.3}} &
  {\color[HTML]{000000} \textit{45.1±0.3}} &
  \multicolumn{1}{c|}{\cellcolor[HTML]{EFEFEF}{\color[HTML]{656565} 25.4±0.4}} &
  {\color[HTML]{000000} \textit{54.8±0.4}} &
  {\color[HTML]{656565} 13.5±0.5} &
  {\color[HTML]{000000} \textit{19.4±0.5}} &
  \multicolumn{1}{c|}{\cellcolor[HTML]{EFEFEF}{\color[HTML]{656565} 45.7±1.0}} &
  {\color[HTML]{000000} \textit{28.0±0.8}} &
  \multicolumn{1}{c|}{\cellcolor[HTML]{EFEFEF}{\color[HTML]{656565} 29.5±0.8}} &
  {\color[HTML]{000000} \textit{38.7±0.4}} &
  {\color[HTML]{656565} 15.5±0.3} \\
\rowcolor[HTML]{EFEFEF} 
{\color[HTML]{000000} MOE-MOSE} &
  {\color[HTML]{000000} \textbf{37.4±0.3}} &
  \multicolumn{1}{c|}{\cellcolor[HTML]{EFEFEF}{\color[HTML]{656565} 34.7±0.3}} &
  {\color[HTML]{000000} \textbf{47.0±0.4}} &
  \multicolumn{1}{c|}{\cellcolor[HTML]{EFEFEF}{\color[HTML]{656565} 23.6±0.4}} &
  {\color[HTML]{000000} \textbf{55.6±0.4}} &
  {\color[HTML]{656565} 12.7±0.4} &
  {\color[HTML]{000000} \textbf{21.4±0.4}} &
  \multicolumn{1}{c|}{\cellcolor[HTML]{EFEFEF}{\color[HTML]{656565} 40.6±0.6}} &
  {\color[HTML]{000000} \textbf{29.8±0.5}} &
  \multicolumn{1}{c|}{\cellcolor[HTML]{EFEFEF}{\color[HTML]{656565} 26.3±0.5}} &
  {\color[HTML]{000000} \textbf{39.3±0.8}} &
  {\color[HTML]{656565} \textit{13.9±0.6}} \\ \bottomrule[1.5pt]
\end{tabular}%
% \vspace{0.3cm}
}
\caption{\textbf{Average Accuracy (ACC) \& Average Forgetting (AF)} across two class-incremental datasets with 3 different memory sizes. We record the mean and standard deviation from 15 random runs. The best and second best results are highlighted using \textbf{bold} and \textit{italic} fonts, respectively. The first 11 rows are online CL algorithms, while the next 7 rows show the performance of offline CL methods. Our proposed method MOSE along with its MOE version are listed in the last two rows, achieving state-of-the-art performance.}
\label{tab:acc_af}
\end{table*}

\section{Experiment}
\label{sec:exp}

\subsection{Experimental Setup}
\label{sec:setup}

\textbf{Dataset.}
We examine two benchmark datasets widely used in OCL, \ie, CIFAR-100 (100 classes)~\cite{cifar} and Tiny-ImageNet (200 classes)~\cite{tinyimagenet}.
Following the setup of previous OCL research~\cite{OCM, OnPro, GSA}, we split CIFAR-100 into 10 disjoint tasks with 10 classes per task (500/100 training/testing samples per class) and split Tiny-ImageNet into 100 disjoint tasks with 2 classes per task (500/50 training/testing samples per class), respectively.

% \vspace{0.2cm}
\noindent
\textbf{Baseline.} We compare our approach with 11 OCL methods (AGEM~\cite{AGEM}, ER~\cite{ER}, MIR~\cite{MIR}, GSS~\cite{GSS}, ASER~\cite{ASER}, ER-AML~\cite{ER-AML}, GDumb~\cite{GDumb}, SCR~\cite{SCR}, OCM~\cite{OCM}, OnPro~\cite{OnPro} and GSA~\cite{GSA}) and 7 offline CL methods (DER++~\cite{DER}, IL2A~\cite{IL2A}, CO$^2$L~\cite{CO2L}, LUCIR~\cite{LUCIR}, CCIL~\cite{CCIL}, BIC~\cite{BIC} and SSIL~\cite{SSIL}). Those offline CL methods are trained with one epoch in each task following the online setting. We use average accuracy (\textbf{ACC}) and average forgetting (\textbf{AF}) of all tasks after learning each task as evaluation metrics~\cite{ASER, DVC} (detailed in Appendix~\ref{sec:ipl_details}). Results of all methods and their publication dates are recorded in Tab.~\ref{tab:acc_af}.

% \vspace{0.2cm}
\noindent
\textbf{Implementation Details.} We use ResNet18~\cite{resnet} with random initialization as the backbone feature extractor $f_{\theta}$, which is composed of 4 convolutional blocks corresponding to the 4 experts $\{E_i\}_{i=1}^4$. We implement the feature alignment modules $\{a_{\omega_i}\}_{i=1}^4$ as convolutional layers (group convolution layers with kernel size $3\times 3$ to shrink feature map size and point convolutional layers with kernel size $1\times 1$ to increase channel dimension). The final expert $E_4$ is feature extractor $f_\theta$ itself, so $a_{\omega_4}$ is the trivial identity function. We use fully-connected layers as projection layers $\{p_{\psi_i}\}_{i=1}^4$ and $\{g_{\phi_i}\}_{i=1}^4$. The original ResNet18 has 11.29M parameters, and our auxiliary networks bring another 1.52M. Still, they are \emph{far less} than methods requiring knowledge distillation with a copy of previous model~\cite{OCM, LWF} and will be \emph{deleted} once the training is completed. An NCM classifier~\cite{SCR, NCM} is used to do class prediction with the feature maps output by $f_{\theta}$ (\ie, the final expert $E_4$). MOSE applies reservoir sampling strategy~\cite{Reservoir} and random memory update. We fix incoming batch $B=10$ and buffer batch size $B^{\mathcal{M}}=64$ for all baselines according to~\cite{ASER, DVC, OCM}. For the two OCL benchmarks, we set memory size $M =1k, 2k, 5k$ and $M=2k, 4k, 10k$, respectively (same as work~\cite{OCM, GSA}). We take Adam as our optimizer, with learning rate $1\times 10 ^{-3}$ and weight decay $1 \times 10 ^{-4}$ for all methods. $\tau$ is $\mathcal{L}_{scl}$ is set to $0.07$ as SCR. Default hyperparameters, as well as code links of all baselines, can be found in Appendix~\ref{sec:baseline_code_hyper}, where their training time is provided to assess their efficiency.

\noindent
\textbf{Data Augmentation.} As discussed in Sec.~\ref{sec:empricial_analysis} we see the positive effect of data augmentation in OCL. Following SimCLR~\cite{simclr} and OCM~\cite{OCM}, we use a transformation operation combining random horizontal flip, random grayscale, and random resized crop for our MOSE as well as all baselines for fair comparison. Specifically, augment both incoming batch $\mathcal{B}^t$ and buffer batch $\mathcal{B}^{\mathcal{M}}$. Notice that SCR~\cite{SCR}, DER++~\cite{DER}, and DVC~\cite{DVC} have implemented their unique augmentation transformations, so we keep the default setup. Global rotation augmentation with inner flip introduced in OCM produces 15 times more training samples and is accepted by later works (GSA~\cite{GSA}, OnPro~\cite{OnPro}). Here we take the inner flip operation to double the training samples, whose impact is discussed in Appendix~\ref{sec:augmentation}.

\subsection{Experimental Result}
\label{sec:main_results}

\textbf{Overall Performance.} 
The overall performance of all methods is summarized in Tab.~\ref{tab:acc_af} with the mean and standard deviation of 15 runs. It can be seen that the proposed MOSE framework achieves substantial average accuracy improvement over all baselines, with a performance lead of more than $5.4\%$ and $5.5\%$ compared with state-of-the-art methods over Split CIFAR-100 and Split Tiny-ImageNet, respectively. Besides, we provide the results of MOSE without using RSD, instead, we calculate the performance of average output logits from each expert, denoted as MOE-MOSE. The results of MOE-MOSE further indicate the superiority of utilizing multiple experts' skills, and it achieves even higher improvement: $7.3\%$ on Split CIFAR-100 and $6.1\%$ on Split Tiny-ImageNet. The outstanding performance on the ACC metric highlights the excellence of our approach to addressing the OCL problem. MOSE's average forgetting over benchmarks is a relatively smaller value than most OCL and offline CL problems. Along with the ACC results, we can conclude that our MOSE gets high test accuracy for each new task when it is under training, and the forgetting does not significantly affect the final performance so ultimately MOSE demonstrates the strongest among all baselines.

\begin{figure}[]
  % \vspace{-0.4cm}
  \centering
  \begin{subfigure}{0.49\linewidth}
    % \fbox{\rule{0pt}{1.5in} \rule{.9\linewidth}{0pt}}
    \includegraphics[width=\linewidth]{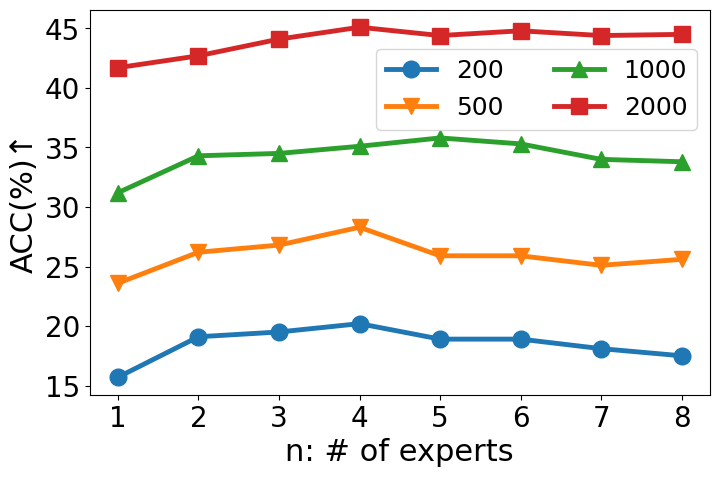}
    \caption{Average End Accuracy ($\uparrow$)}
    \label{fig:acc}
  \end{subfigure}
  \begin{subfigure}{0.49\linewidth}
    % \fbox{\rule{0pt}{1.5in} \rule{.9\linewidth}{0pt}}
    \includegraphics[width=\linewidth]{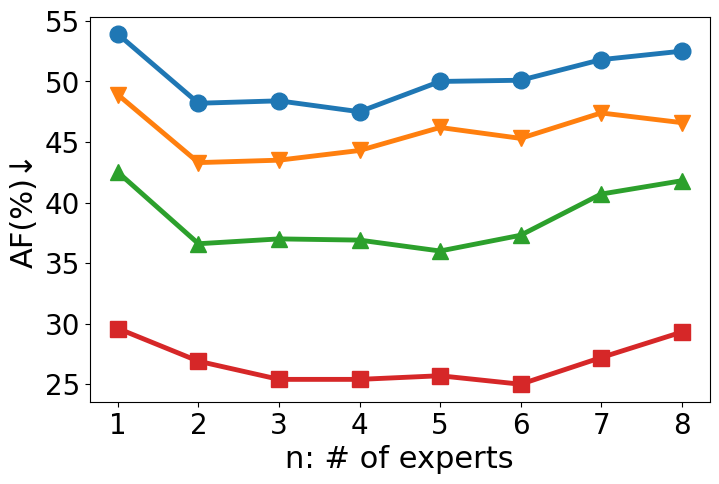}
    \caption{Average End Forgetting ($\downarrow$)}
    \label{fig:af}
  \end{subfigure} \\
  \caption{\textbf{Different Number of Experts.} We divide the ResNet18 backbone into a few components according to its block-wise structure, evaluated under four different memory buffer sizes.}
  \label{fig:num_of_exp}
  % \vspace{-0.9cm}
\end{figure}

\noindent
\textbf{Number of Experts.}
The selection $n=4$ naturally comes from the architecture of the ResNet, as discussed in Sec.~\ref{sec:setup}. In Fig.~\ref{fig:num_of_exp}, we show ACC and AF for different $n$ and buffer sizes over Split-CIFAR100. $n=4$ balances both sides, w/o increasing too much complexity or cost.

\noindent
\textbf{Small Buffers.} For scalability and transferability concerns, we present experiment results with small buffers in Tab.~\ref{tab:small}. Here we show the ACC for both datasets except we utilize a 10-task configuration for Split Tiny-ImageNet due to the subpar performance of all baselines in the 100-task version when constrained by limited memory capacities, rendering them inadequate for comparison.

\begin{figure}[]
  % \vspace{-0.1cm}
  \centering
  \begin{subfigure}{0.49\linewidth}
    % \fbox{\rule{0pt}{1.5in} \rule{.9\linewidth}{0pt}}
    \includegraphics[width=\linewidth]{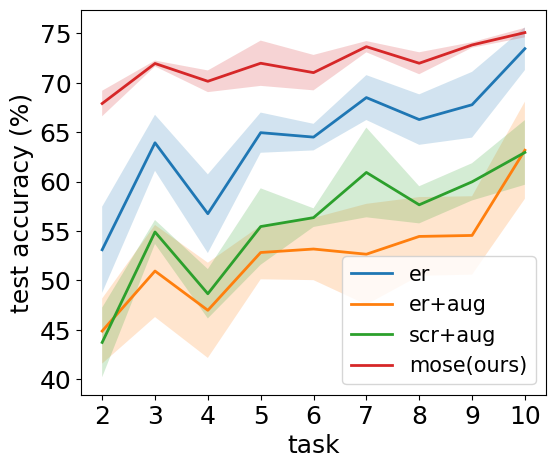}
    \caption{Test accuracy of new task}
    \label{fig:test}
  \end{subfigure}
  \begin{subfigure}{0.49\linewidth}
    % \fbox{\rule{0pt}{1.5in} \rule{.9\linewidth}{0pt}}
    \includegraphics[width=\linewidth]{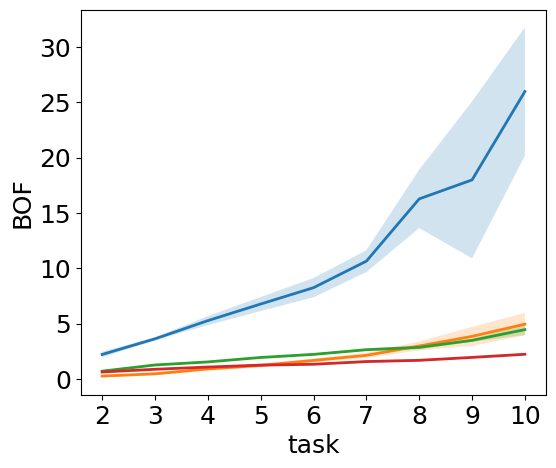}
    \caption{Average BOF of old tasks}
    \label{fig:bof}
  \end{subfigure} \\
  \caption{\textbf{Overfitting-Underfitting Test}. These two subfigures exhibit \textbf{(a)} the test accuracy of each new task $t$ when it is trained; and \textbf{(b)} the average BOF value of old tasks after learning each task.
  }
  \label{fig:over-under}
  % \vspace{-0.1cm}
\end{figure}

\noindent
\textbf{Addressing Overfitting-Underfitting.}
In response to the overfitting-underfitting dilemma discussed in Sec.~\ref{sec:prelinminary}, we record the new task test accuracy and average BOF (Eq.\ref{eq:bof}) over old tasks in Fig.\ref{fig:over-under}, following the same experiment setup in Sec.\ref{sec:empricial_analysis}. These two subfigures demonstrate that MOSE learns each new task well with higher new task test accuracy (avoid underfitting of new task) and mitigates buffer overfitting for old tasks (lower BOF value throughout the entire training period), therefore it resolves both sides of this dilemma pretty well, which explains the origin of its advantages in OCL benchmarks.

%\vspace{-0.1cm}

\begin{table}[]
    % \renewcommand\arraystretch{0.4}
    % \vspace{-0.4cm}
    \centering
    \resizebox{1.0\linewidth}{!}{
\begin{tabular}{@{}r|cc|cc@{}}
\toprule[2pt]
\multicolumn{1}{c|}{\textbf{Dataset}} & \multicolumn{2}{c|}{\textbf{Split CIFAR-100 (10 tasks)}} & \multicolumn{2}{c}{\textbf{Split Tiny-ImageNet(10 tasks)}} \\ \midrule
\multicolumn{1}{c|}{Memory Size} & 200      & 500      & 500      & 1K     \\ \midrule
OCM(2022)                         & 12.2±0.4 & 19.7±0.5 & 7.3±0.5  & 10.5±0.6 \\
OnPro(2023)                       & 14.1±0.9 & 21.5±1.4 & 7.2±0.4  & 10.2±0.3 \\
GSA(2023)                         & 14.9±0.3 & 22.9±0.2 & 10.4±0.3 & 14.8±0.2 \\
MOSE(ours)                                  & \textbf{20.2±0.5}        & \textbf{28.3±0.7}       & \textbf{15.2±0.7}         & \textbf{20.2±0.9}        \\ \bottomrule[1.5pt]
\end{tabular}
    }
    % \vspace{-0.3cm}
    \caption{\textbf{Small Buffer Scenario.} ACC recorded for 15 runs.}
    \label{tab:small}
    % \vspace{-0.2cm}
\end{table}

\begin{table}[]
\renewcommand\arraystretch{0.9}
\centering
  % \vspace{-0.1cm}
\resizebox{0.4\textwidth}{!}{%
\begin{tabular}{@{}l|ccc|l@{}}
\toprule[2pt]
\textbf{Loss} & $\mathcal{L}_{ce}$ & $\mathcal{L}_{scl}$ & $\mathcal{L}_{\text{RSD}}$ & \textbf{ACC (\%)} \\ \midrule
\multirow{3}{*}{w/o MLS} & last layer   &              &              & 36.89          \\
                         &              & last layer   &              & 39.36          \\
                         & last layer   & last layer   &              & 45.00          \\ \midrule \midrule
\multirow{4}{*}{w/ MLS}  & $\checkmark$ &              &              & 40.24          \\
                         &              & $\checkmark$ &              & 46.21          \\
                         & $\checkmark$ & $\checkmark$ &              & 52.29          \\
                         & $\checkmark$ & $\checkmark$ & $\checkmark$ & \textbf{54.57} \\ \bottomrule[1.5pt]
\end{tabular}%
}
\caption{\textbf{Ablation Study} with impact of individual building bricks of MOSE over Split CIFAR-100 dataset and $M=5000$. Records are averaged over 15 random runs. The best result is in \textbf{bold} font.}
\label{tab:ablation}
% \vspace{-4em}
  % \vspace{-0.2cm}
\end{table}
%\vspace{-0.1cm}

\subsection{Ablation Study}
\label{sec:ablation}

For each building block of our design, multi-level experts with supervision $\mathcal{L}_{\text{MLS}}$ and reverse self-distillation $\mathcal{L}_{\text{RSD}}$, we experiment on dataset Split CIFAR-100 with memory size $M=5k$ (see Tab.~\ref{tab:ablation}). We observe a consistent increase in average accuracy with the inclusion of each component. This validates the efficacy of individual modules and their collaborative contribution to the success of MOSE. 

Also, Tab.~\ref{tab:moe_acc} verifies the effectiveness of RSD: the mission of it -- concentrates experts' capabilities onto the final expert -- is accomplished. Simultaneously, with the network divided into multiple experts, they each excel in different tasks, forming a \textit{diagonal pattern} where the maximum values for each task are presented in the first half of Tab.~\ref{tab:moe_acc}. On the one hand, this indicates differences in the learned encoding abilities between network hierarchies, with shallower layers possibly possessing better generalization, preserving better accuracy on the initial task by avoiding drastic feature changes during subsequent training. On the other hand, the features obtained in the final layer may be more advanced and semantic, leading to good performance on newer tasks. RSD helps the final expert $E_4$ learn tasks in which it performs poorly, showing the efficacy of knowledge transfer within the learner network in OCL.

\section{Conclusion and Discussion}
\label{sec:conclusion}

This paper attributes the distinction between OCL and regular CL to the particular challenge termed overfitting-underfitting dilemma for sequentially arriving tasks. The proposed method MOSE utilizes multi-level latent experts and self-distillation to integrate different skills across network hierarchies, achieving appropriate convergence in each task. Extensive experiments demonstrate the remarkable superiority of MOSE over state-of-the-art baselines and its effectiveness in addressing the overfitting-underfitting dilemma. 
We expect this to facilitate the understanding of OCL and inspire subsequent research of latent expertise for CL-related problems.
In the future, we will combine MOSE with more backbone architectures and different supervision signals to further unleash its potential.

\section{Acknowledgement}
\label{sec:ack}
% \textcolor{red}{\td add more ack msg.}
This project is supported by the Tsinghua-Peking Center for Life Sciences (to Y.Z.), the National Natural Science Foundation of China (20211710187, 32021002), and the Wafer-level Silicon Photonic Interconnect On-chip Computing System (2022YFB2804100). L.W. is supported by the Postdoctoral Fellowship Program of CPSF under Grant Number GZB20230350 and the Shuimu Tsinghua Scholar.

\clearpage
{
    \small
    \bibliographystyle{ieeenat_fullname}
    \bibliography{main}
}

\clearpage
\setcounter{page}{1}
\maketitlesupplementary

\section{Implementation Details}
\label{sec:ipl_details}

Here's the overall algorithm of our proposed method MOSE in a standard OCL training scenario.

\begin{algorithm}[h]
    %\caption{MOSE framework for OCL}
    \caption{Training Algorithm of MOSE}
    \begin{algorithmic}[1]\label{alg:mose}
    \REQUIRE ~\\
        $\mathcal{D} = \{\mathcal{D}_t\}_{t\leq T}$: training dataset \\
        Network $F$ with its experts $\{E_i\}_{i\leq n}$ \\
        $\text{Aug}(\cdot)$: augmentation transformation \\
    \ENSURE Multi-level supervision of sequential experts and reverse self-distillation in the OCL network
    \FORALL{task $t\leq T$}
        \FORALL{incoming batch: $\mathcal{B}^t\sim \mathcal{D}_t$}
            \STATE Memory retrieval: $\mathcal{B}^\mathcal{M}\sim \mathcal{M}$, $\mathcal{B} = \mathcal{B}^t \cup \mathcal{B}^{\mathcal{M}}$
            \STATE Augmentation: $\tilde{\mathcal{B}} = \mathcal{B} \cup \text{Aug}(\mathcal{B})$
            \STATE Calculate $\mathcal{L}_{\text{MLS}}$ and $\mathcal{L}_{\text{RSD}}$ over $\tilde{\mathcal{B}}$
            \STATE Update $F$, $\{E_i\}_{i\leq n}$ with $\mathcal{L}_{\text{MOSE}} =  \mathcal{L}_{\text{MLS}} + \mathcal{L}_{\text{RSD}}$
            \STATE Memory update: $\mathcal{M}\leftarrow \mathcal{M}, \mathcal{B}^t$
        \ENDFOR    
    \ENDFOR
    \end{algorithmic}
\end{algorithm}

\begin{table}[h]
    % \renewcommand\arraystretch{0.5}
    % \vspace{-0.4cm}
    \centering
    \resizebox{1.0\linewidth}{!}{
\begin{tabular}{@{}r|c|c|c|c@{}}
\toprule[2pt]
\multicolumn{1}{c|}{\textbf{Buffer}} & \textbf{200}      & \textbf{500}      & \textbf{1000}     & \textbf{2000}     \\ \midrule
SD     & 11.9±0.8 & 22.0±0.8 & 30.6±0.7 & 41.9±0.4 \\
RSD (ours)                           & \textbf{20.2±0.5} & \textbf{28.3±0.7} & \textbf{35.1±0.4} & \textbf{45.1±0.3} \\ \bottomrule[1.5pt]
\end{tabular}
    }
    % \vspace{-0.3cm}
    \caption{\textbf{SD \textit{v.s.} RSD.} ACC over Split CIFAR-100 for four different memory buffer sizes.}
    \label{tab:sd_vs_rsd}
    % \vspace{-0.6cm}
\end{table}

\begin{table}[h]
    % \renewcommand\arraystretch{0.5}
    % \vspace{-0.4cm}
    \centering
    \resizebox{1.0\linewidth}{!}{
\begin{tabular}{@{}c|ccc|cc@{}}
\toprule[2pt]
\textbf{Model} & \textbf{Time} & \textbf{\# Params} & \textbf{\#FLOPs} & \textbf{ACC} $\uparrow$ & \textbf{AF} $\downarrow$ \\ \midrule
SCR+$\mathcal{L}_{ce}$     & 8.8min  & 11.29M & 558.55M & 22.3±0.4 & 45.8±0.2 \\
w/ MLS     & 17.3min & 12.81M & 586.78M & 31.6±0.8 & 39.1±1.2 \\
w/ MLS,RSD & 17.5min & 12.81M & 590.64M & 35.1±0.4 & 36.9±0.3 \\ \bottomrule[1.5pt]
\end{tabular}
    }
    % \vspace{-0.3cm}
    \caption{\textbf{Training Cost of Components.} Recorded for Split CIFAR-100 with $M=5k$.}
    \label{tab:cost}
    % \vspace{-0.4cm}
\end{table}

\subsection{Network Architecture}
\label{sec:net_details}

\begin{table*}[p]
\renewcommand\arraystretch{1.5}
\resizebox{\textwidth}{!}{%
\begin{tabular}{@{}r|cccc@{}}
\toprule[2pt]
\textbf{Expert}  & \textbf{$a_{\omega_1}$} & \textbf{$a_{\omega_2}$}         & \textbf{$a_{\omega_3}$}         & \textbf{$a_{\omega_4}$}       \\ \midrule
input shape& (16, 32, 32)                        & (128, 16, 16)          & (256, 8, 8)            & (512, 4, 4)            \\ \midrule
\multirow{2}{*}{composition} & $
\left[
\begin{aligned}
& \text{Conv3x3($C_{\text{in}}$, $C_{\text{in}}$, s=2, g=$C_{\text{in}}$)} \\
& \text{Conv1x1($C_{\text{in}}$, $C_{\text{in}}$, s=1, g=1)} \\
& \text{BatchNorm($C_{\text{in}}$)} \\
& \text{ReLU()} \\
& \text{Conv3x3($C_{\text{in}}$, $C_{\text{in}}$, s=1, g=$C_{\text{in}}$)} \\
& \text{Conv1x1($C_{\text{in}}$, $C_{\text{in}} \times 2$, s=1, g=1)} \\
& \text{BatchNorm($C_{\text{in}} \times 2$)} \\
& \text{ReLU()}
\end{aligned}
\right] \times 3
$    & $
\left[
\begin{aligned}
& \text{Conv3x3($C_{\text{in}}$, $C_{\text{in}}$, s=2, g=$C_{\text{in}}$)} \\
& \text{Conv1x1($C_{\text{in}}$, $C_{\text{in}}$, s=1, g=1)} \\
& \text{BatchNorm($C_{\text{in}}$)} \\
& \text{ReLU()} \\
& \text{Conv3x3($C_{\text{in}}$, $C_{\text{in}}$, s=1, g=$C_{\text{in}}$)} \\
& \text{Conv1x1($C_{\text{in}}$, $C_{\text{in}} \times 2$, s=1, g=1)} \\
& \text{BatchNorm($C_{\text{in}} \times 2$)} \\
& \text{ReLU()}
\end{aligned}
\right] \times 2
$    & $
\left[
\begin{aligned}
& \text{Conv3x3($C_{\text{in}}$, $C_{\text{in}}$, s=2, g=$C_{\text{in}}$)} \\
& \text{Conv1x1($C_{\text{in}}$, $C_{\text{in}}$, s=1, g=1)} \\
& \text{BatchNorm($C_{\text{in}}$)} \\
& \text{ReLU()} \\
& \text{Conv3x3($C_{\text{in}}$, $C_{\text{in}}$, s=1, g=$C_{\text{in}}$)} \\
& \text{Conv1x1($C_{\text{in}}$, $C_{\text{in}} \times 2$, s=1, g=1)} \\
& \text{BatchNorm($C_{\text{in}} \times 2$)} \\
& \text{ReLU()}
\end{aligned}
\right] \times 1
$    & Identity()    \\
                 & +AdaptiveAvgPool2d(1,1)              & +AdaptiveAvgPool2d(1,1) & +AdaptiveAvgPool2d(1,1) & +AdaptiveAvgPool2d(1,1) \\ \midrule
output dimension               & 512 & 512 & 512 & 512 \\ \midrule
\# of parameters               & 268800 & 254976 & 202752 & 0 \\ \bottomrule[1.5pt]
\end{tabular}%
}
\caption{\textbf{Composition of Alignment Modules.} Each alignment module contains consecutive similar blocks to down-sample the maps' sizes and increase their channel dimensions. \textit{Conv3x3} and \textit{Conv1x1} are convolutional layers with kernel size $(3,3)$ and $(1,1)$, respectively. $s$ stands for stride and $g$ means number of groups. \textit{BatchNorm}, \textit{ReLU}, and \textit{AdaptiveAvgPool2d} are common network layers: batch normalization, ReLU activation, and adaptive average 2D pooling layers. \textit{Identity()} function outputs the input feature map directly.}
\label{tab:alignment_modules}
\end{table*}

Our model is based on the architecture of a full-width ResNet18~\cite{resnet} as mentioned in Sec.~\ref{sec:setup}, where the original embedding dimension of feature vector output by the feature extractor $f_\theta$ is $d=512$. The feature maps produced by each block are of different sizes in shape: (channel, height, width). Therefore the feature alignment modules $\{a_{\omega_i}\}_{i=1}^4$ are designed to further encode those feature maps into the same dimension space with as little computational cost as possible. Here we list the details of the necessary components of $\{a_{\omega_i}\}_{i=1}^4$ in Tab.~\ref{tab:alignment_modules}.

As for projection heads $\{p_{\psi_i}\}_{i=1}^4$ and $\{g_{\phi_i}\}_{i=1}^4$, we use fully connected layers $\text{Linear}(512, l)$ and $\text{Linear}(512, |\mathcal{C}|)$ to project the feature vector into a new embedding space of dimension $l=128$ (as SCR~\cite{SCR} and OCM~\cite{OCM}) and the logit space with dimension equal to the number of image classes $|\mathcal{C}|$. Alignment modules and projection heads will all be dropped once the network's continual learning procedure is accomplished. At the testing phase, we only calculate the feature means for each class in the feature space with dimension $d=512$ and use the NCM classifier~\cite{NCM, SCR}. There is no other additional storage cost.

\subsection{Evaluation Metrics}
\label{sec:evaluation_metrics}

We first train all continual learners task-wise with corresponding training samples, then test them with test samples for all classes when training is complete. We use two commonly used metrics Average Accuracy (\textbf{ACC}) and Average Forgetting (\textbf{AF}) to evaluate their performance on benchmarks following~\cite{mai2022OCLreview, Albin2023OCLreview, OCM, ASER, DVC}:
\begin{equation}
    \begin{split}
        \text{ACC} &= \frac{1}{T} \sum_{t=1}^T \text{acc}_{t, T}, \\
        \text{AF} &= \frac{1}{T-1}\sum_{t=1}^{T-1} \left(\max_{i\leq T-1}{\text{acc}_{t, i}} - \text{acc}_{t, T}\right),
    \end{split}
\label{eq:acc_af}
\end{equation}
where $a_{i, j}$ is the test accuracy of task $i$ when the model has been trained on task from 1 to $j$, and $T$ is the total number of tasks. ACC represents the final remaining skills for all incrementally learned tasks, which is ultimately the optimization goal of a continual learner, and AF shows how the CL algorithm resists the catastrophic forgetting issue.

\section{Additional Ablation Studies}

In this section, we present the additional results of the ablation study to discuss the effect of our choices of each algorithm component on the final OCL performance (average task accuracy). We take Split CIFAR-100 with 10 tasks as the benchmark, and all tests are conducted over different setups with memory buffer sizes $M=1000$, 2000, and 5000 for 15 different random runs.

\subsection{RSD Variations}
\label{sec:rsd_variations}

In Sec.~\ref{sec:reverse_sd}, we use the task-wise final test accuracy ($a_{i, T}$) of each expert with or without the RSD loss within Tab.\ref{tab:moe_acc}. It shows that RSD successfully transfers useful knowledge from different teacher experts $E_{i\leq 3}$ to the final predictor $E_4=F$. Compared to traditional self-distillation (SD)~\cite{SKD}, RSD enhances the final predictor on purpose. We show the difference of ACC between SD and RSD in Tab.~\ref{tab:sd_vs_rsd}, indicating the effectiveness of our artificial modification to distillation direction.

\begin{figure*}[p]
  \centering
  \begin{subfigure}{0.33\linewidth}
    \includegraphics[width=\linewidth]{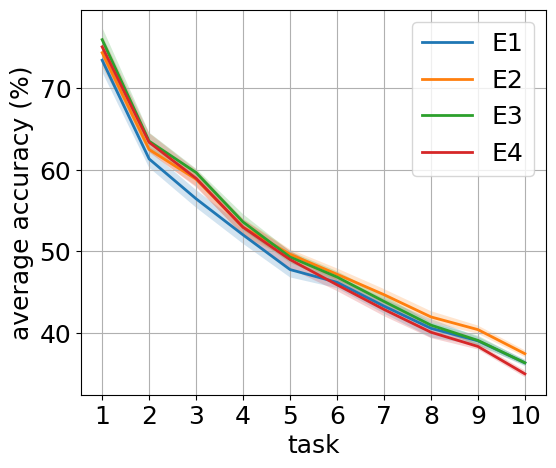}
    \caption{Average Accuracy at $M=1k$}
  \end{subfigure}
  \begin{subfigure}{0.33\linewidth}
    \includegraphics[width=\linewidth]{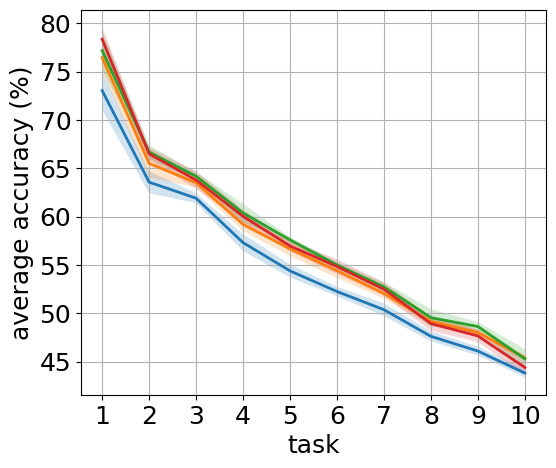}
    \caption{Average Accuracy at $M=2k$}
  \end{subfigure}
  \begin{subfigure}{0.33\linewidth}
    \includegraphics[width=\linewidth]{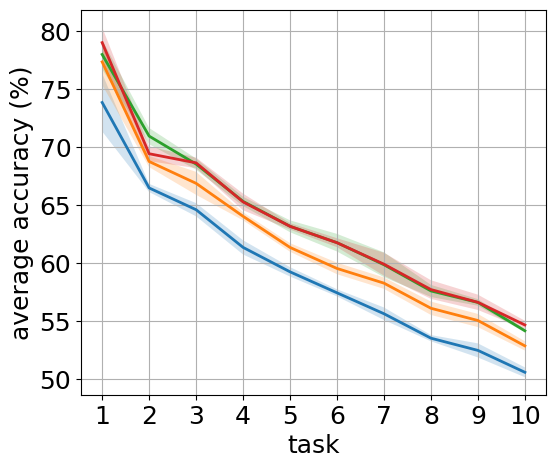}
    \caption{Average Accuracy at $M=5k$}
  \end{subfigure} \\
  \begin{subfigure}{0.33\linewidth}
    \includegraphics[width=\linewidth]{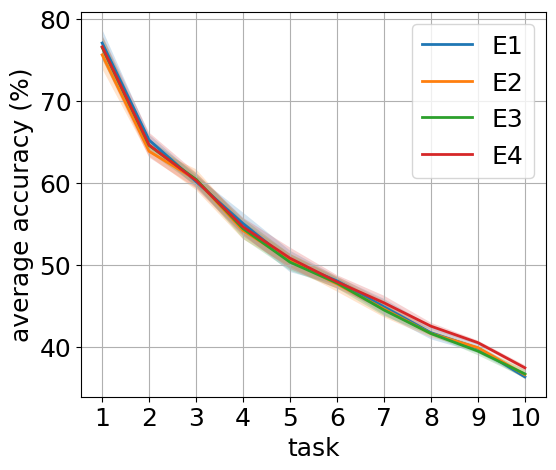}
    \caption{MOE Average Accuracy at $M=1k$}
  \end{subfigure}
  \begin{subfigure}{0.33\linewidth}
    \includegraphics[width=\linewidth]{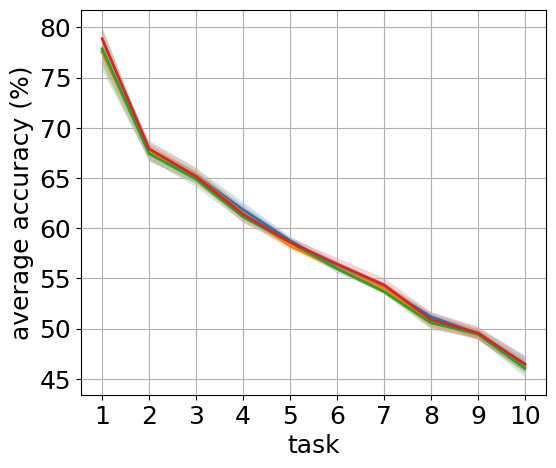}
    \caption{MOE Average Accuracy at $M=2k$}
  \end{subfigure}
  \begin{subfigure}{0.33\linewidth}
    \includegraphics[width=\linewidth]{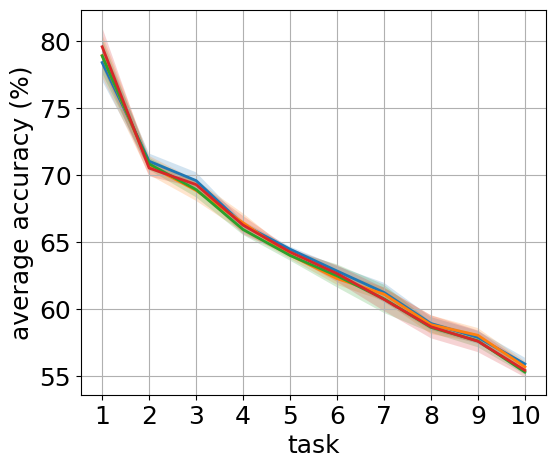}
    \caption{MOE Average Accuracy at $M=5k$}
  \end{subfigure} \\
  \caption{\textbf{Average Accuracy with Different Student Expert.} Here presents the average test accuracy $\mathbb{E}_{i\leq t} a_{i, t}$ at different task $t$ during training. E1, E2, E3, and E4 denote the student experts used in our proposed RSD. \textbf{(a)}, \textbf{(b)} and \textbf{(c)} are accuracy results of corresponding student expert; \textbf{(d)}, \textbf{(e)} and \textbf{(f)} are accuracy results of their MOE version, which is the accuracy of averaged output logits across all experts.
  }  
  \label{fig:rsd_avg}
\end{figure*}

\begin{figure*}[p]
  \centering
  \begin{subfigure}{0.33\linewidth}
    \includegraphics[width=\linewidth]{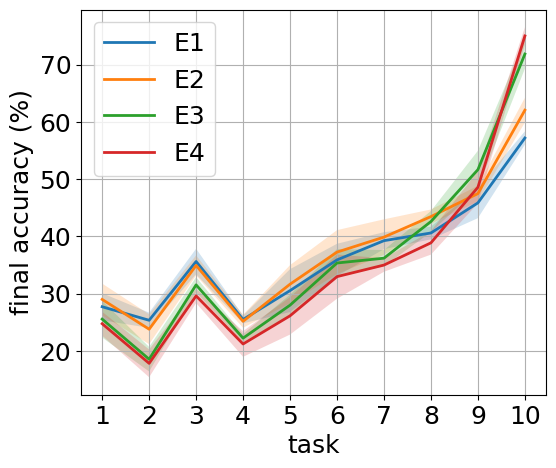}
    \caption{Final Accuracy at $M=1k$}
  \end{subfigure}
  \begin{subfigure}{0.33\linewidth}
    \includegraphics[width=\linewidth]{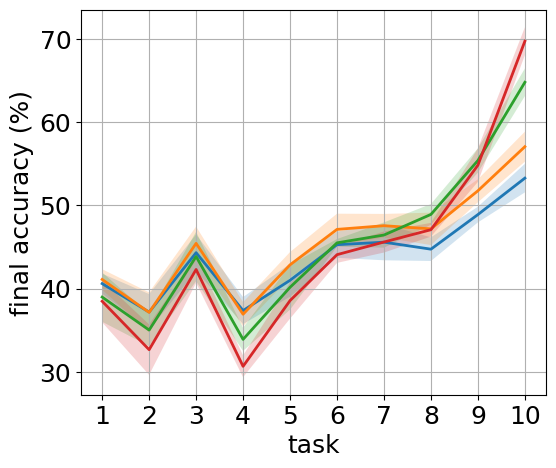}
    \caption{Final Accuracy at $M=2k$}
  \end{subfigure}
  \begin{subfigure}{0.33\linewidth}
    \includegraphics[width=\linewidth]{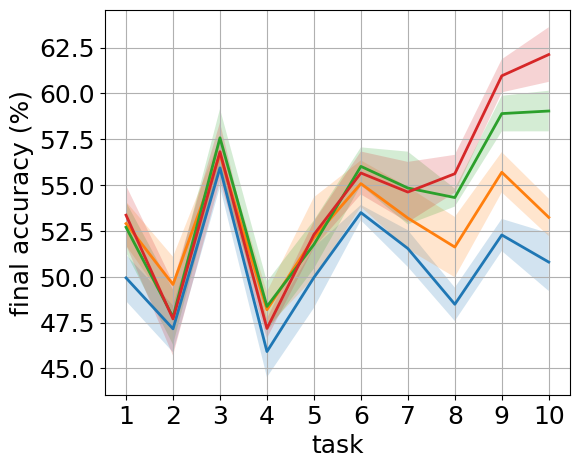}
    \caption{Final Accuracy at $M=5k$}
  \end{subfigure} \\
  \begin{subfigure}{0.33\linewidth}
    \includegraphics[width=\linewidth]{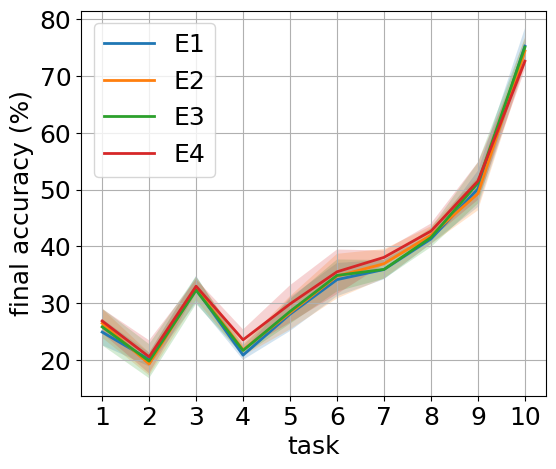}
    \caption{MOE Final Accuracy at $M=1k$}
  \end{subfigure}
  \begin{subfigure}{0.33\linewidth}
    \includegraphics[width=\linewidth]{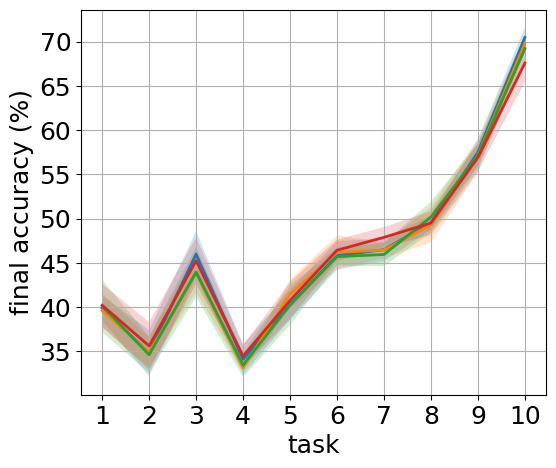}
    \caption{MOE Final Accuracy at $M=2k$}
  \end{subfigure}
  \begin{subfigure}{0.33\linewidth}
    \includegraphics[width=\linewidth]{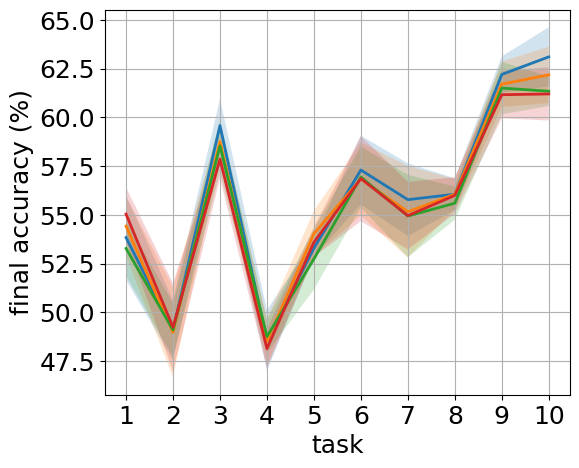}
    \caption{MOE Final Accuracy at $M=5k$}
  \end{subfigure} \\
  \caption{\textbf{Final Task-wise Accuracy with Different Student Expert.} Here presents the final task-wise test accuracy $a_{t, T}$ of different task $t$ after OCL training is complete. E1, E2, E3, and E4 denote the student experts used in our proposed RSD. \textbf{(a)}, \textbf{(b)} and \textbf{(c)} are accuracy results of corresponding student expert; \textbf{(d)}, \textbf{(e)} and \textbf{(f)} are accuracy results of their MOE version.
  }  
  \label{fig:rsd_final}
\end{figure*}

We further test RSD with different experts being the student expert who learns from others. Average accuracy and $\mathbb{E}_{i\leq t} a_{i, t}$ and final task-wise test accuracy $a_{t, T}$ of different task $t$ are depicted in Fig.~\ref{fig:rsd_avg} and Fig.~\ref{fig:rsd_final}. They suggest that: (1) choosing a different expert as the student does not significantly affect the MOE performance; (2) with a larger memory buffer size $M$, variation grows and experts at deeper stages exhibit greater potential of learning from other experts. Detailed results data of $a_{t, T}$ in Tab.~\ref{tab:rsd_final} make these points more clear.

\begin{table*}[p]
\resizebox{\textwidth}{!}{%
\begin{tabular}{@{}ll|llllllllll|l@{}}
\toprule[2pt]
\multicolumn{2}{c|}{\textbf{Experts}} &
  \textbf{task 1} &
  \textbf{task 2} &
  \textbf{task 3} &
  \textbf{task 4} &
  \textbf{task 5} &
  \textbf{task 6} &
  \textbf{task 7} &
  \textbf{task 8} &
  \textbf{task 9} &
  \textbf{task 10} &
  \textbf{Average} \\ \midrule
\multicolumn{1}{l|}{\multirow{4}{*}{$M=1k$}} &
  $E_1$ &
  27.72 &
  \textbf{25.36} &
  \textbf{35.62} &
  \textbf{25.50} &
  30.54 &
  35.90 &
  39.26 &
  40.60 &
  45.86 &
  57.18 &
  36.35 \\
\multicolumn{1}{l|}{} &
  $E_2$ &
  \textbf{29.00} &
  23.80 &
  34.94 &
  25.18 &
  \textbf{31.66} &
  \textbf{37.26} &
  \textbf{39.88} &
  \textbf{43.46} &
  47.48 &
  62.04 &
  \textbf{37.47} \\
\multicolumn{1}{l|}{} &
  $E_3$ &
  25.56 &
  18.56 &
  31.54 &
  22.24 &
  28.00 &
  35.36 &
  36.20 &
  42.60 &
  \textbf{51.66} &
  71.84 &
  36.36 \\
\multicolumn{1}{l|}{} &
  $E_4$ &
  24.76 &
  17.84 &
  29.62 &
  21.22 &
  26.14 &
  32.98 &
  35.00 &
  38.88 &
  48.58 &
  \textbf{75.00} &
  35.00 \\ \midrule \midrule
\multicolumn{1}{l|}{\multirow{4}{*}{$M=2k$}} &
  $E_1$ &
  40.60 &
  \textbf{37.18} &
  44.32 &
  \textbf{37.36} &
  41.00 &
  45.28 &
  45.58 &
  44.74 &
  48.90 &
  53.28 &
  43.82 \\
\multicolumn{1}{l|}{} &
  $E_2$ &
  \textbf{41.10} &
  37.14 &
  \textbf{45.44} &
  36.92 &
  \textbf{42.86} &
  \textbf{47.12} &
  \textbf{47.56} &
  47.18 &
  51.76 &
  57.06 &
  \textbf{45.41} \\
\multicolumn{1}{l|}{} &
  $E_3$ &
  38.98 &
  35.02 &
  43.86 &
  33.90 &
  40.16 &
  45.50 &
  46.44 &
  \textbf{48.92} &
  \textbf{55.40} &
  64.80 &
  45.30 \\
\multicolumn{1}{l|}{} &
  $E_4$ &
  38.48 &
  32.66 &
  42.30 &
  30.66 &
  38.56 &
  44.08 &
  45.60 &
  47.08 &
  54.84 &
  \textbf{69.74} &
  44.40 \\ \midrule \midrule
\multicolumn{1}{l|}{\multirow{4}{*}{$M=5k$}} &
  $E_1$ &
  49.94 &
  47.16 &
  55.94 &
  45.92 &
  49.96 &
  53.50 &
  51.54 &
  48.50 &
  52.28 &
  50.80 &
  50.55 \\
\multicolumn{1}{l|}{} &
  $E_2$ &
  52.90 &
  \textbf{49.58} &
  56.66 &
  48.20 &
  52.14 &
  55.08 &
  53.20 &
  51.62 &
  55.70 &
  53.24 &
  52.83 \\
\multicolumn{1}{l|}{} &
  $E_3$ &
  52.70 &
  47.80 &
  \textbf{57.58} &
  \textbf{48.38} &
  51.76 &
  \textbf{56.02} &
  \textbf{54.84} &
  54.32 &
  58.90 &
  59.04 &
  54.13 \\
\multicolumn{1}{l|}{} &
  $E_4$ &
  \textbf{53.36} &
  47.70 &
  56.82 &
  47.18 &
  \textbf{52.30} &
  55.66 &
  54.62 &
  \textbf{55.62} &
  \textbf{60.96} &
  \textbf{62.12} &
  \textbf{54.63} \\ \bottomrule[1.5pt]
\end{tabular}%
}
\caption{\textbf{Final Task-wise Accuracy with Different Student Expert}. Here presents the final task-wise test accuracy $a_{t, T}$ of different task $t$ after OCL training is complete. $E_1$, $E_2$, $E_3$, and $E_4$ denote the student experts used in our proposed RSD.}
\label{tab:rsd_final}
\end{table*}

\subsection{Data Augmentation}
\label{sec:augmentation}

% \hw{1. Use SimCLR augmentation or not; 2. Use rotation or not. (compared with different baselines)}

\begin{figure*}[]
  \centering
  \begin{subfigure}{0.33\linewidth}
    \includegraphics[width=\linewidth]{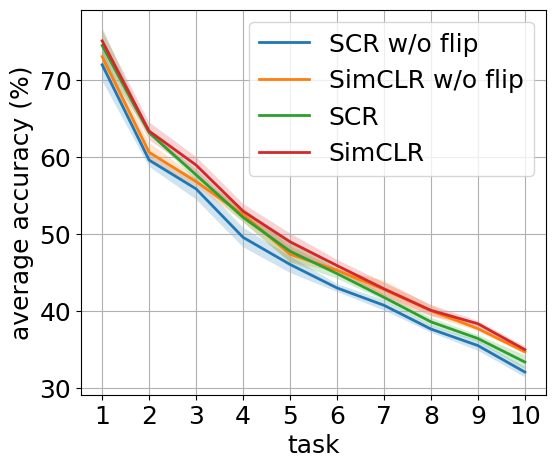}
    \caption{Average Accuracy at $M=1k$}
  \end{subfigure}
  \begin{subfigure}{0.33\linewidth}
    \includegraphics[width=\linewidth]{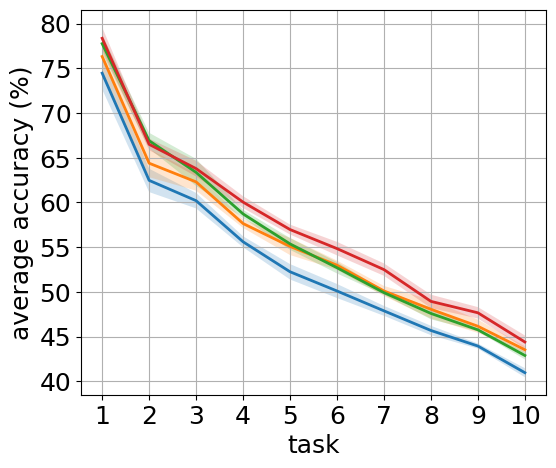}
    \caption{Average Accuracy at $M=2k$}
  \end{subfigure}
  \begin{subfigure}{0.33\linewidth}
    \includegraphics[width=\linewidth]{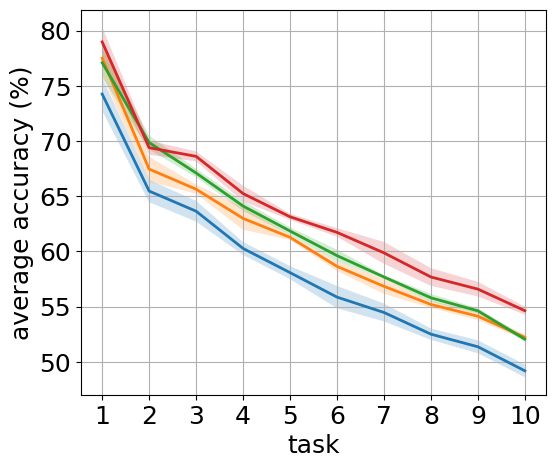}
    \caption{Average Accuracy at $M=5k$}
  \end{subfigure} \\
  \begin{subfigure}{0.33\linewidth}
    \includegraphics[width=\linewidth]{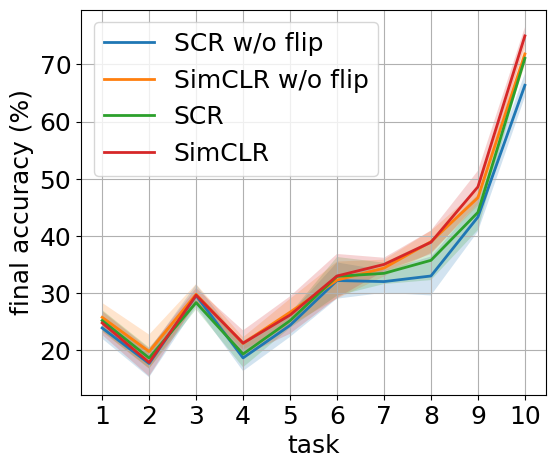}
    \caption{Final Accuracy at $M=1k$}
  \end{subfigure}
  \begin{subfigure}{0.33\linewidth}
    \includegraphics[width=\linewidth]{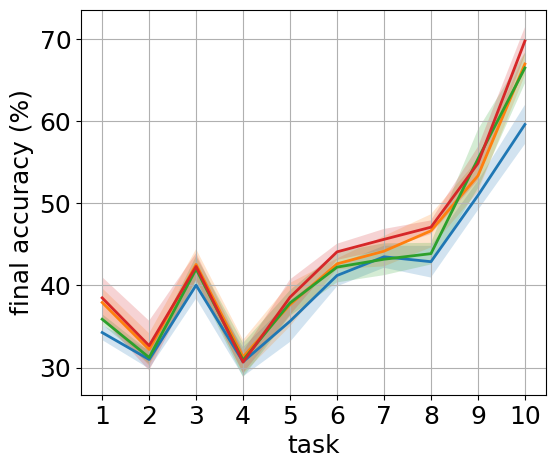}
    \caption{Final Accuracy at $M=2k$}
  \end{subfigure}
  \begin{subfigure}{0.33\linewidth}
    \includegraphics[width=\linewidth]{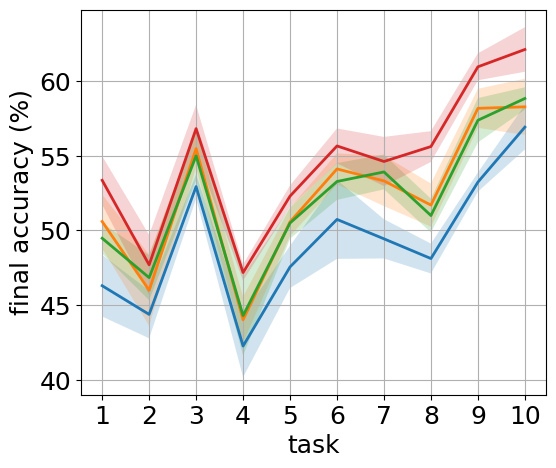}
    \caption{Final Accuracy at $M=5k$}
  \end{subfigure} \\
  \caption{\textbf{Data augmentation.} Here presents the average test accuracy $\mathbb{E}_{i\leq t} a_{i, t}$ and final task-wise accuracy $a_{t, T}$ at different task $t$ during training. Four different training setups, where we use \textit{SCR} or \textit{SimCLR} augmentation combination and whether inner flip is introduced, are tested with three memory buffer sizes. \textbf{(a)}, \textbf{(b)} and \textbf{(c)} are average test accuracy results; \textbf{(d)}, \textbf{(e)} and \textbf{(f)} are final task-wise accuracy results.
  }  
  \label{fig:da}
\end{figure*}

As described in Sec.\ref{sec:setup}, we use a transformation operation combining random horizontal flip, random grayscale, and random resized crop, following SimCLR~\cite{simclr} and OCM~\cite{OCM} (we denote this data augmentation combination as \textit{SimCLR}). Our proposed method MOSE also utilizes the contrastive loss $\mathcal{L}_{scl}$ of SCR~\cite{SCR}, where a different data augmentation combination is used: random resized crop, random horizontal flip, color jitter, and random grayscale (we denote it as \textit{SCR}). For situations with both types of augmentation and whether inner flip operation (proposed in OCM~\cite{OCM}) is used to double the training samples, we conducted a contrast experiment whose results are recorded in Fig.~\ref{fig:da} (average accuracy and final task-wise accuracy) and Tab.~\ref{tab:da_final} (final task-wise accuracy).

We conclude from the above tests that, the choice of augmentation indeed affects the performance of our method MOSE. The usage of a more complex and diverse augmentation combination does not guarantee better accuracy. For example, the insertion of color jitter or inner flip results in quite different task-wise performance. However, increasing the number of training samples through augmentation is validated to be an effective approach. This type of improvement remains consistent across different memory buffer sizes.

The inner flip operation rotates half of the image vertically. Thus there are in total $2\times 2=4$ different possible flipped versions of one image. On top of that, OCM~\cite{OCM} and OnPro~\cite{OnPro} also utilize global rotation and create $4\times 4=16$ augmented versions of each single image. In order not to further increase the number of computations and GPU memory usage of training so that the comparison with other methods is fair, we choose to use inner flip only to double the training samples. Even so, the final performance improvement brought by MOSE is already evident.

\begin{table*}[p]
\resizebox{\textwidth}{!}{%
\begin{tabular}{@{}ll|llllllllll|l@{}}
\toprule[2pt]
\multicolumn{2}{c|}{\textbf{Experts}} &
  \textbf{task 1} &
  \textbf{task 2} &
  \textbf{task 3} &
  \textbf{task 4} &
  \textbf{task 5} &
  \textbf{task 6} &
  \textbf{task 7} &
  \textbf{task 8} &
  \textbf{task 9} &
  \textbf{task 10} &
  \textbf{Average} \\ \midrule
\multicolumn{1}{l|}{\multirow{4}{*}{$M=1k$}} &
  SCR w/o flip &
  23.88 &
  17.66 &
  29.56 &
  18.64 &
  24.34 &
  32.18 &
  32.00 &
  32.98 &
  43.30 &
  66.36 &
  32.09 \\
\multicolumn{1}{l|}{} &
  SimCLR w/o flip &
  \textbf{25.74} &
  \textbf{19.74} &
  \textbf{29.78} &
  21.18 &
  \textbf{26.64} &
  32.26 &
  34.36 &
  \textbf{38.98} &
  46.68 &
  71.84 &
  34.72 \\
\multicolumn{1}{l|}{} &
  SCR &
  25.18 &
  18.62 &
  28.32 &
  19.32 &
  25.16 &
  32.92 &
  33.44 &
  35.72 &
  44.12 &
  71.08 &
  33.39 \\
\multicolumn{1}{l|}{} &
  SimCLR &
  24.76 &
  17.84 &
  29.62 &
  \textbf{21.22} &
  26.14 &
  \textbf{32.98} &
  \textbf{35.00} &
  38.88 &
  \textbf{48.58} &
  \textbf{75.00} &
  \textbf{35.00} \\ \midrule \midrule
\multicolumn{1}{l|}{\multirow{4}{*}{$M=2k$}} &
  SCR w/o flip &
  34.26 &
  30.98 &
  40.02 &
  30.70 &
  35.64 &
  41.20 &
  43.48 &
  42.88 &
  50.94 &
  59.58 &
  40.97 \\
\multicolumn{1}{l|}{} &
  SimCLR w/o flip &
  37.92 &
  32.20 &
  \textbf{42.54} &
  \textbf{31.30} &
  37.66 &
  42.62 &
  44.16 &
  46.64 &
  53.34 &
  66.94 &
  43.53 \\
\multicolumn{1}{l|}{} &
  SCR &
  35.88 &
  31.22 &
  41.98 &
  30.90 &
  37.86 &
  42.22 &
  43.18 &
  43.86 &
  \textbf{55.38} &
  66.46 &
  42.89 \\
\multicolumn{1}{l|}{} &
  SimCLR &
  \textbf{38.48} &
  \textbf{32.66} &
  42.30 &
  30.66 &
  \textbf{38.56} &
  \textbf{44.08} &
  \textbf{45.60} &
  \textbf{47.08} &
  54.84 &
  \textbf{69.74} &
  \textbf{44.40} \\ \midrule \midrule
\multicolumn{1}{l|}{\multirow{4}{*}{$M=5k$}} &
  SCR w/o flip &
  46.30 &
  44.38 &
  52.94 &
  42.26 &
  47.56 &
  50.74 &
  49.44 &
  48.12 &
  53.24 &
  56.92 &
  49.19 \\
\multicolumn{1}{l|}{} &
  SimCLR w/o flip &
  50.60 &
  46.00 &
  55.48 &
  44.02 &
  50.58 &
  54.12 &
  53.32 &
  51.70 &
  58.18 &
  58.28 &
  52.23 \\
\multicolumn{1}{l|}{} &
  SCR &
  49.48 &
  46.84 &
  55.00 &
  44.30 &
  50.50 &
  53.28 &
  53.92 &
  51.00 &
  57.38 &
  58.84 &
  52.05 \\
\multicolumn{1}{l|}{} &
  SimCLR &
  \textbf{53.36} &
  \textbf{47.70} &
  \textbf{56.82} &
  \textbf{47.18} &
  \textbf{52.30} &
  \textbf{55.66} &
  \textbf{54.62} &
  \textbf{55.62} &
  \textbf{60.96} &
  \textbf{62.12} &
  \textbf{54.63} \\ \bottomrule[1.5pt]
\end{tabular}%
}
\caption{\textbf{Data augmentation.} Here presents the final task-wise accuracy $a_{t, T}$ at different task $t$ during training. Four different training setups, where we use \textit{SCR} or \textit{SimCLR} augmentation combination and whether inner flip is introduced, are tested with three memory sizes.
}
\label{tab:da_final}
\end{table*}

\subsection{NCM Classifiers}

% \hw{Whether to use NCM classifier}

\begin{figure*}[]
  \centering
  \begin{subfigure}{0.33\linewidth}
    \includegraphics[width=\linewidth]{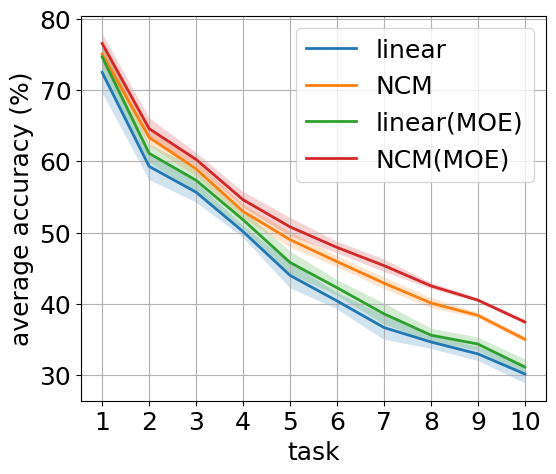}
    \caption{Average Accuracy at $M=1k$}
  \end{subfigure}
  \begin{subfigure}{0.33\linewidth}
    \includegraphics[width=\linewidth]{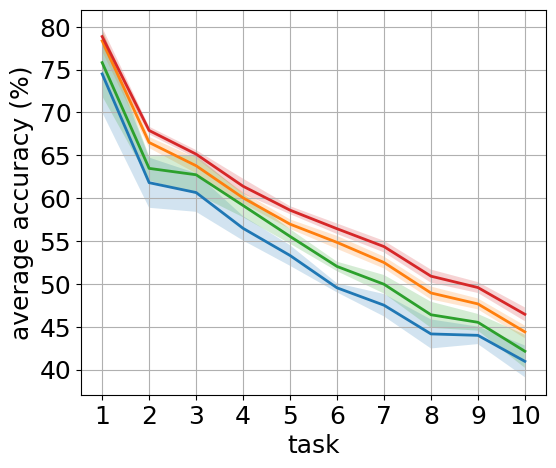}
    \caption{Average Accuracy at $M=2k$}
  \end{subfigure}
  \begin{subfigure}{0.33\linewidth}
    \includegraphics[width=\linewidth]{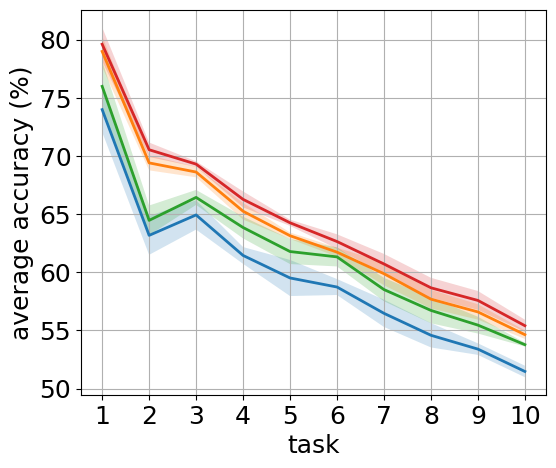}
    \caption{Average Accuracy at $M=5k$}
  \end{subfigure} \\
  \begin{subfigure}{0.33\linewidth}
    \includegraphics[width=\linewidth]{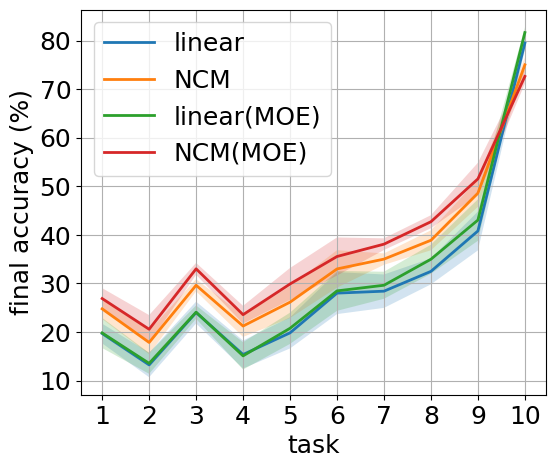}
    \caption{Final Accuracy at $M=1k$}
  \end{subfigure}
  \begin{subfigure}{0.33\linewidth}
    \includegraphics[width=\linewidth]{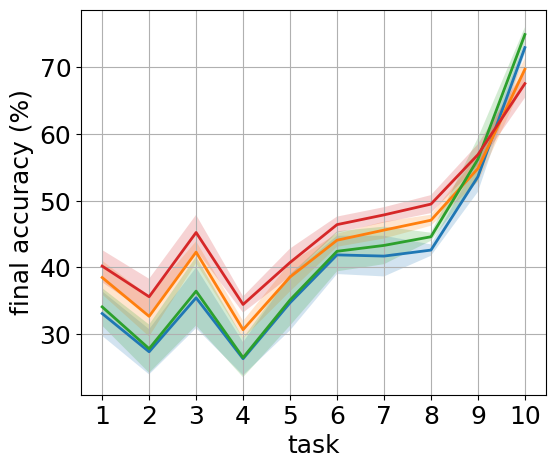}
    \caption{Final Accuracy at $M=2k$}
  \end{subfigure}
  \begin{subfigure}{0.33\linewidth}
    \includegraphics[width=\linewidth]{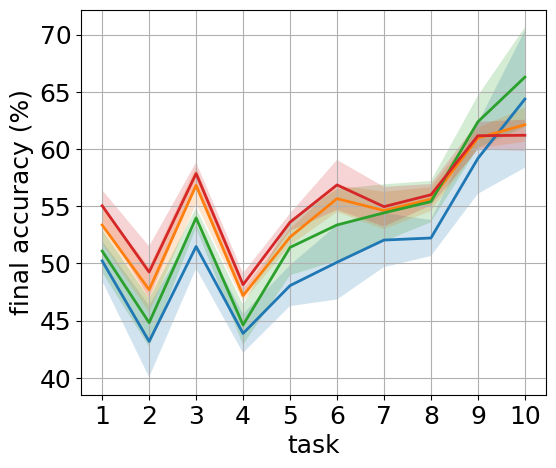}
    \caption{Final Accuracy at $M=5k$}
  \end{subfigure} \\
  \caption{\textbf{NCM or Linear Classifier.} Here presents the average test accuracy $\mathbb{E}_{i\leq t} a_{i, t}$ and final task-wise accuracy $a_{t, T}$ at different task $t$ during training. Four different training setups, where we use \textit{NCM} or \textit{linear} classifier along with its MOE versions, are tested with three memory buffer sizes. \textbf{(a)}, \textbf{(b)} and \textbf{(c)} are average test accuracy results; \textbf{(d)}, \textbf{(e)} and \textbf{(f)} are final task-wise accuracy results.
  }  
  \label{fig:ncm}
\end{figure*}

For each expert $\{E_{i}\}_{i\leq 4}$ we choose to use the NCM classifier to output the final prediction. Compared to the traditional linear output head, the NCM classifier is based on feature representation and is constrained by the memory buffer size. Here we show the results of using these two types of classifiers (along with the MOE version) under our MOSE framework in Fig.~\ref{fig:ncm} (average accuracy and final task-wise accuracy) and Tab.~\ref{tab:ncm_final} (final task-wise accuracy).

From the above tests, we see that these two types of classifiers present similar inter-task behavior: the last trained task has a significantly larger test accuracy than the previous ones. The representation-based NCM classifier generally outperforms the linear classifier, indicating its superiority within our proposed framework which relies on multi-level feature representation learning.

\begin{table*}[p]
\resizebox{\textwidth}{!}{%
\begin{tabular}{@{}ll|llllllllll|l@{}}
\toprule[2pt]
\multicolumn{2}{c|}{\textbf{Experts}} &
  \textbf{task 1} &
  \textbf{task 2} &
  \textbf{task 3} &
  \textbf{task 4} &
  \textbf{task 5} &
  \textbf{task 6} &
  \textbf{task 7} &
  \textbf{task 8} &
  \textbf{task 9} &
  \textbf{task 10} &
  \textbf{Average} \\ \midrule
\multicolumn{1}{l|}{\multirow{4}{*}{$M=1k$}} &
  linear &
  19.68 &
  13.22 &
  24.00 &
  15.36 &
  19.82 &
  28.00 &
  28.40 &
  32.48 &
  40.80 &
  79.50 &
  30.13 \\
\multicolumn{1}{l|}{} &
  NCM &
  24.76 &
  17.84 &
  29.62 &
  21.22 &
  26.14 &
  32.98 &
  35.00 &
  38.88 &
  48.58 &
  75.00 &
  35.00 \\
\multicolumn{1}{l|}{} &
  linear(MOE) &
  19.80 &
  13.54 &
  24.08 &
  15.08 &
  20.76 &
  28.46 &
  29.62 &
  34.98 &
  43.04 &
  \textbf{81.64} &
  31.10 \\
\multicolumn{1}{l|}{} &
  NCM(MOE) &
  \textbf{26.88} &
  \textbf{20.56} &
  \textbf{33.00} &
  \textbf{23.56} &
  \textbf{29.88} &
  \textbf{35.54} &
  \textbf{38.08} &
  \textbf{42.70} &
  \textbf{51.52} &
  72.62 &
  \textbf{37.43} \\ \midrule \midrule
\multicolumn{1}{l|}{\multirow{4}{*}{$M=2k$}} &
  linear &
  33.08 &
  27.34 &
  35.44 &
  26.30 &
  34.70 &
  41.88 &
  41.70 &
  42.60 &
  53.62 &
  72.98 &
  40.96 \\
\multicolumn{1}{l|}{} &
  NCM &
  38.48 &
  32.66 &
  42.30 &
  30.66 &
  38.56 &
  44.08 &
  45.60 &
  47.08 &
  54.84 &
  69.74 &
  44.40 \\
\multicolumn{1}{l|}{} &
  linear(MOE) &
  34.08 &
  27.80 &
  36.44 &
  26.46 &
  35.12 &
  42.42 &
  43.30 &
  44.60 &
  56.24 &
  \textbf{74.94} &
  42.14 \\
\multicolumn{1}{l|}{} &
  NCM(MOE) &
  \textbf{40.20} &
  \textbf{35.60} &
  \textbf{45.24} &
  \textbf{34.44} &
  \textbf{40.74} &
  \textbf{46.42} &
  \textbf{47.88} &
  \textbf{49.50} &
  \textbf{56.92} &
  67.58 &
  \textbf{46.45} \\ \midrule \midrule
\multicolumn{1}{l|}{\multirow{4}{*}{$M=5k$}} &
  linear &
  50.22 &
  43.18 &
  51.48 &
  43.88 &
  48.06 &
  50.10 &
  52.04 &
  52.22 &
  59.18 &
  64.36 &
  51.47 \\
\multicolumn{1}{l|}{} &
  NCM &
  53.36 &
  \textbf{47.70} &
  56.82 &
  47.18 &
  52.30 &
  55.66 &
  54.62 &
  55.62 &
  60.96 &
  62.12 &
  54.63 \\
\multicolumn{1}{l|}{} &
  linear(MOE) &
  51.08 &
  44.82 &
  53.98 &
  44.62 &
  51.40 &
  53.36 &
  54.42 &
  55.40 &
  \textbf{62.38} &
  \textbf{66.28} &
  53.77 \\
\multicolumn{1}{l|}{} &
  NCM(MOE) &
  \textbf{55.04} &
  49.24 &
  \textbf{57.86} &
  \textbf{48.14} &
  \textbf{53.60} &
  \textbf{56.86} &
  \textbf{54.96} &
  \textbf{56.00} &
  61.16 &
  61.20 &
  \textbf{55.41} \\ \bottomrule[1.5pt]
\end{tabular}%
}
\caption{\textbf{Data augmentation.} Here presents the final task-wise accuracy $a_{t, T}$ at different task $t$ during training. Four different training setups, where we use \textit{NCM} or \textit{linear} classifier along with its MOE versions, are tested with three memory buffer sizes.
}
\label{tab:ncm_final}
\end{table*}

\section{Baseline Codes and Efficiency}
\label{sec:baseline_code_hyper}

% \hw{1.List the implementation details of each baseline; 2.Source code links; 3.Running time comparison.}
For all baselines tested in Sec.~\ref{sec:exp}, the common hyperparameters are fixed to give a fair comparison, including batch size $B=10$, memory buffer size $B^{\mathcal{M}}=64$ and random seed $0$. For other algorithm-specific setups, we keep the default implementation based on their source codes in Tab.~\ref{tab:baseline_source_codes} and refer to the training details of OCM~\cite{OCM}.

\begin{table*}[p]
\centering
\resizebox{0.6\textwidth}{!}{%
\begin{tabular}{@{}r|l@{}}
\toprule[2pt]
\textbf{Baseline} & \multicolumn{1}{c}{\textbf{Source Code Links}}                                                         \\ \midrule
AGEM              & \href{https://github.com/facebookresearch/agem}{https://github.com/facebookresearch/agem}             \\
ER and MIR   & \href{https://github.com/optimass/Maximally_Interfered_Retrieval}{https://github.com/optimass/Maximally\_Interfered\_Retrieval}           \\
GSS          & \href{https://github.com/rahafaljundi/Gradient-based-Sample-Selection}{https://github.com/rahafaljundi/Gradient-based-Sample-Selection} \\
ASER and SCR & \href{https://github.com/RaptorMai/online-continual-learning}{https://github.com/RaptorMai/online-continual-learning}                   \\
ER-AML            & \href{https://github.com/pclucas14/AML}{https://github.com/pclucas14/AML}                             \\
GDumb             & \href{https://github.com/drimpossible/GDumb}{https://github.com/drimpossible/GDumb}                   \\
OCM               & \href{https://github.com/gydpku/OCM}{https://github.com/gydpku/OCM}                                   \\
OnPro             & \href{https://github.com/weilllllls/OnPro}{https://github.com/weilllllls/OnPro}                       \\
GSA               & \href{https://github.com/gydpku/GSA}{https://github.com/gydpku/GSA}                                   \\
DER++             & \href{https://github.com/aimagelab/mammoth}{https://github.com/aimagelab/mammoth}                     \\
IL2A              & \href{https://github.com/Impression2805/IL2A}{https://github.com/Impression2805/IL2A}                 \\
Co$^2$L           & \href{https://github.com/chaht01/Co2L}{https://github.com/chaht01/Co2L}                               \\
LUCIR        & \href{https://github.com/hshustc/CVPR19_Incremental_Learning}{https://github.com/hshustc/CVPR19\_Incremental\_Learning}                   \\
CCIL              & \href{https://github.com/sud0301/essentials_for_CIL}{https://github.com/sud0301/essentials\_for\_CIL}   \\
BiC               & \href{https://github.com/sairin1202/BIC}{https://github.com/sairin1202/BIC}                           \\
SSIL              & \href{https://github.com/hongjoon0805/SS-IL-Official}{https://github.com/hongjoon0805/SS-IL-Official} \\ \bottomrule[1.5pt]
\end{tabular}%
}
\caption{\textbf{Baseline Source Code Links.} We list the official source links of all tested baseline algorithms.}
\label{tab:baseline_source_codes}
\end{table*}

\subsection{Running Time Comparison.}

\begin{figure*}[p]
  \centering
    \includegraphics[width=0.9\linewidth]{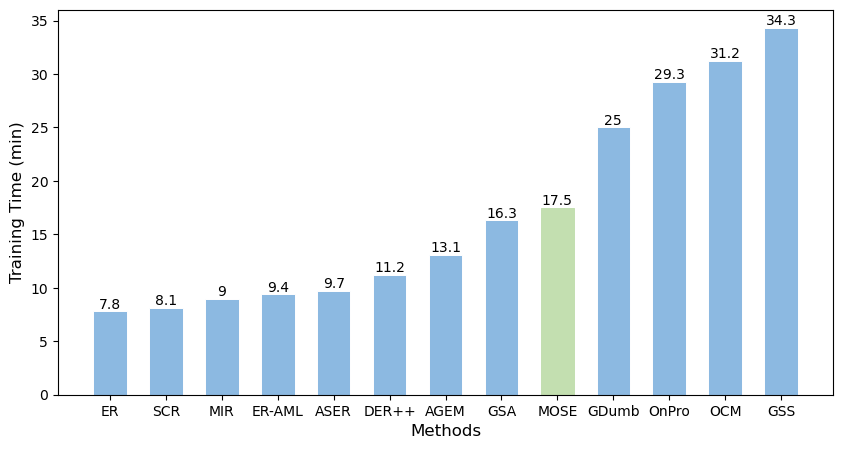}
  \caption{\textbf{Training Time (min).} Here we show the training time of all tested OCL methods over the dataset Split CIFAR-100 with memory size $M=5000$. Time for our proposed method MOSE is colored in \textcolor[RGB]{195,223,176}{green} for better visual presentation.
  }  
  \label{fig:training_time}
\end{figure*}

We run all of our experiments on an NVIDIA RTX3090-Turbo GPU. Here to display the training efficiency, we record the training time over the Split CIFAR-100 dataset for all OCL baselines including MOSE in Fig.~\ref{fig:training_time}. Notice that the introduction of data augmentation also increases the training time. Tab.~\ref{tab:cost} demonstrates the training cost and computational complexity for components of MOSE.

Our proposed MOSE achieves a relatively efficient result compared to other baselines with a training time of around 15 minutes over the Split CIFAR-100 dataset (50000 images in total). The calculation of expert-wise loss doubles the training time of each image compared to ER~\cite{ER} and SCR~\cite{SCR} which have the same supervision loss as MOSE.

\section{Overfitting-Underfitting Dilemma}

% \hw{Test with different MEM size and expert-wise (with the final expert being the student or not)}
To further investigate the multi-level expert design in addressing the proposed overfitting-underfitting dilemma, we record the new task accuracy $a_{t,t}$ for each expert with or without using RSD (take the final expert $E_4$ as the student), as well as average BOF value of old tasks (see Eq.~\ref{eq:bof}). 

\begin{figure*}[]
  \centering
  \begin{subfigure}{0.30\linewidth}
    \includegraphics[width=\linewidth]{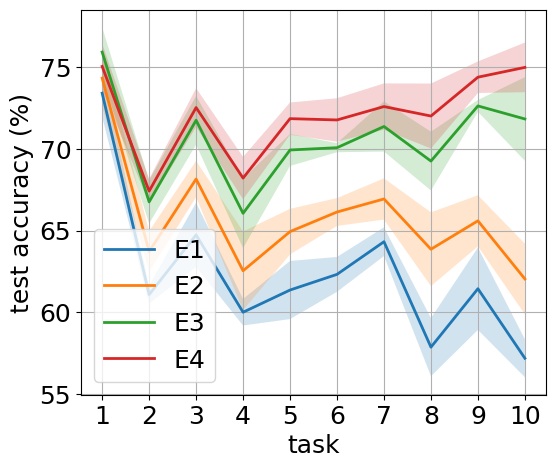}
    \caption{New Task Accuracy at $M=1k$}
  \end{subfigure}
  \begin{subfigure}{0.30\linewidth}
    \includegraphics[width=\linewidth]{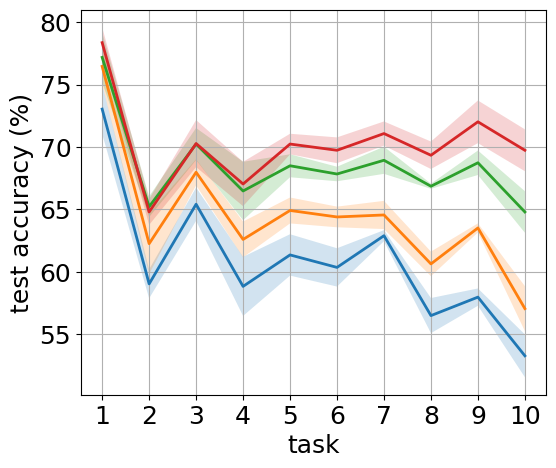}
    \caption{New Task Accuracy at $M=2k$}
  \end{subfigure}
  \begin{subfigure}{0.30\linewidth}
    \includegraphics[width=\linewidth]{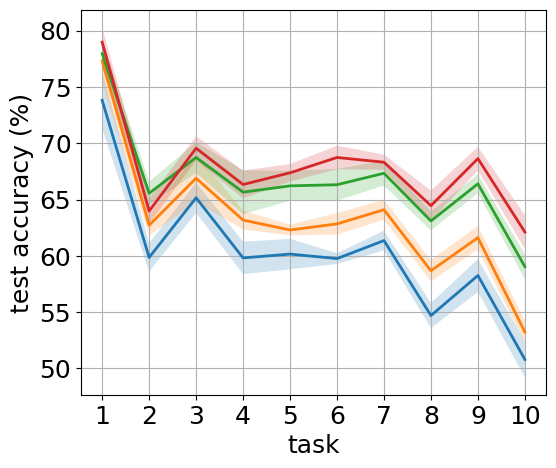}
    \caption{New Task Accuracy at $M=5k$}
  \end{subfigure} \\
  \begin{subfigure}{0.30\linewidth}
    \includegraphics[width=\linewidth]{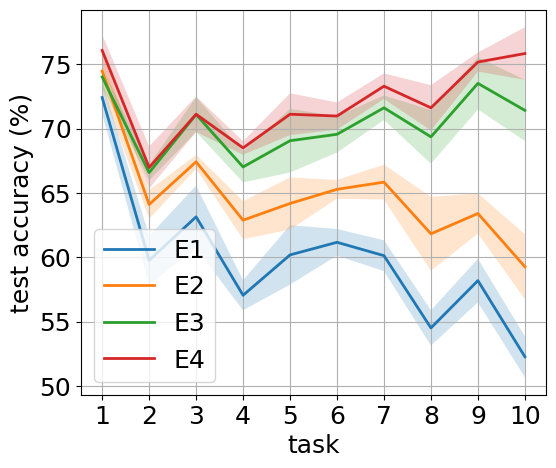}
    \caption{New Task Accuracy with RSD at $M=1k$}
  \end{subfigure}
  \begin{subfigure}{0.30\linewidth}
    \includegraphics[width=\linewidth]{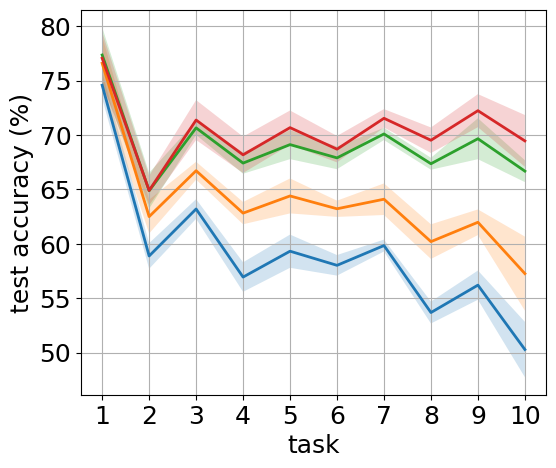}
    \caption{New Task Accuracy with RSD at $M=2k$}
  \end{subfigure}
  \begin{subfigure}{0.30\linewidth}
    \includegraphics[width=\linewidth]{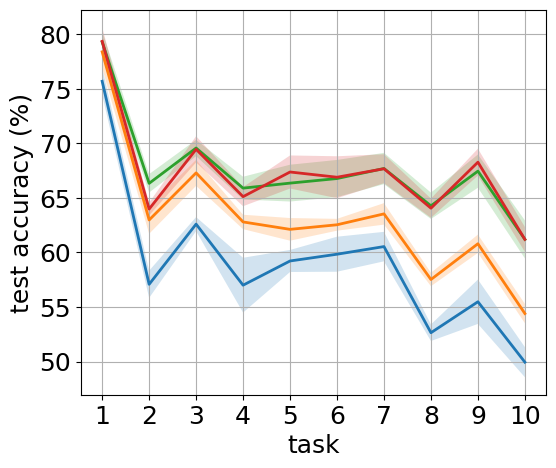}
    \caption{New Task Accuracy with RSD at $M=5k$}
  \end{subfigure} \\
  \caption{\textbf{New Task Accuracy with or without RSD.} Here presents the new task accuracy $a_{t, t}$ at different task $t$ during training. The results of all experts with three memory buffer sizes are depicted. \textbf{(a)}, \textbf{(b)} and \textbf{(c)}: no RSD loss $\mathcal{L}_{\text{RSD}}$; \textbf{(d)}, \textbf{(e)} and \textbf{(f)}: with RSD loss $\mathcal{L}_{\text{RSD}}$.
  }  
  \label{fig:new_expert}
\end{figure*}

\begin{figure*}[]
  \centering
  \begin{subfigure}{0.30\linewidth}
    \includegraphics[width=\linewidth]{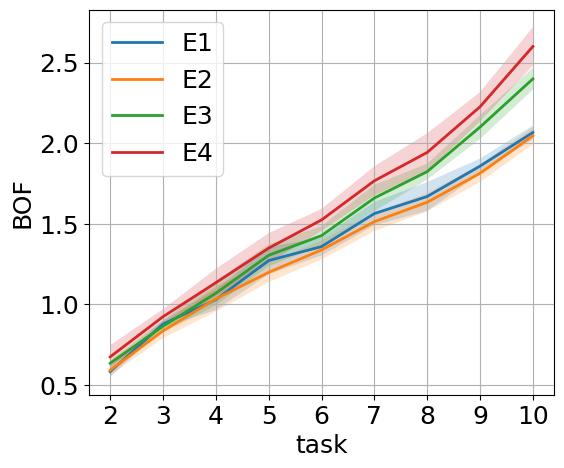}
    \caption{BOF at $M=1k$}
  \end{subfigure}
  \begin{subfigure}{0.30\linewidth}
    \includegraphics[width=\linewidth]{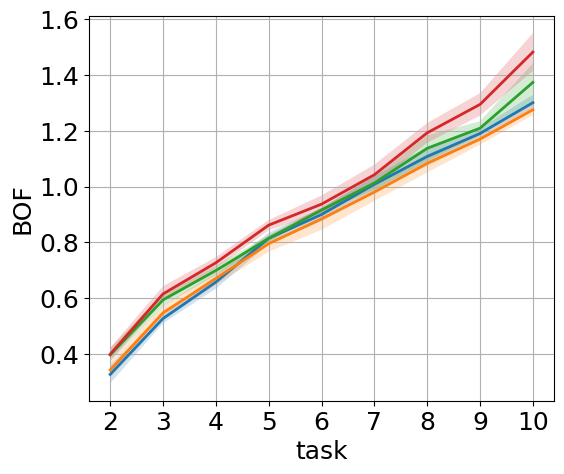}
    \caption{BOF at $M=2k$}
  \end{subfigure}
  \begin{subfigure}{0.30\linewidth}
    \includegraphics[width=\linewidth]{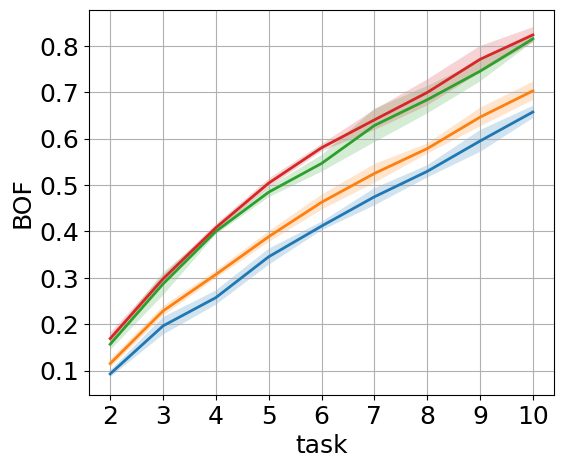}
    \caption{BOF at $M=5k$}
  \end{subfigure} \\
  \begin{subfigure}{0.30\linewidth}
    \includegraphics[width=\linewidth]{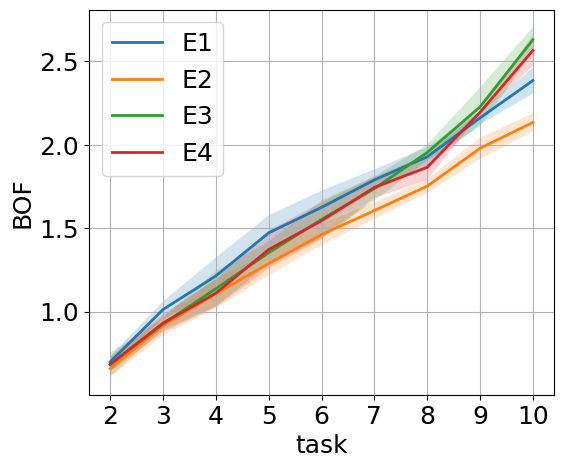}
    \caption{BOF with RSD at $M=1k$}
  \end{subfigure}
  \begin{subfigure}{0.30\linewidth}
    \includegraphics[width=\linewidth]{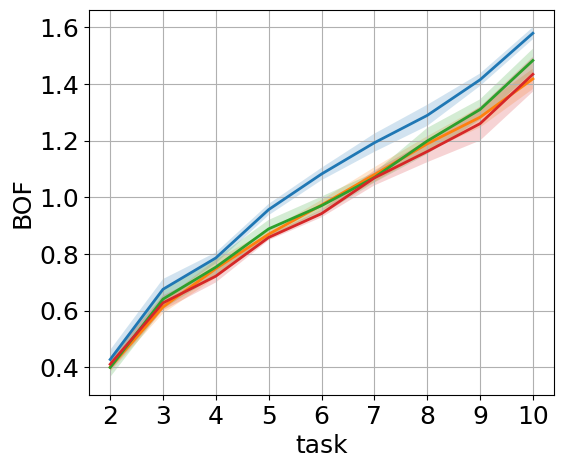}
    \caption{BOF with RSD at $M=2k$}
  \end{subfigure}
  \begin{subfigure}{0.30\linewidth}
    \includegraphics[width=\linewidth]{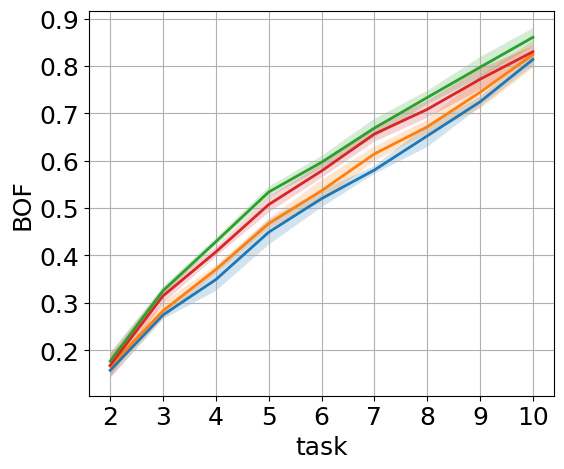}
    \caption{BOF with RSD at $M=5k$}
  \end{subfigure} \\
  \caption{\textbf{Average BOF with or without RSD.} Here presents the average BOF value (Eq.~\ref{eq:bof}) at different task $t$ during training. The results of all experts with three memory buffer sizes are depicted. \textbf{(a)}, \textbf{(b)} and \textbf{(c)}: no RSD loss $\mathcal{L}_{\text{RSD}}$; \textbf{(d)}, \textbf{(e)} and \textbf{(f)}: with RSD loss $\mathcal{L}_{\text{RSD}}$.
  }  
  \label{fig:bof_expert}
\end{figure*}

Accuracy $a_{t,t}$ represents the performance of learning each new incoming task, and BOF indicates how well the buffer overfitting issue is dealt with (lower is better). All tests are conducted over the dataset Split CIFAR-100 with three different memory sizes for 15 random runs. Based on results in Fig.~\ref{fig:new_expert} and Fig.~\ref{fig:bof_expert}, we explain the effectiveness of MOSE in solving the overfitting-underfitting dilemma as follows:

\begin{enumerate}
    \item From Fig.~\ref{fig:new_expert}, it is clear that experts have different performances in learning new tasks: deeper ones ($E_{3}$ and $E_{4}$) generally learn better than experts ($E_1$ and $E_2$) from shallower layers. The RSD loss has no noticeable effect on $E_4$ the last expert's learning of the new task, which means its ability to avoid the underfitting problem of the new task is preserved.
    \item From Fig.~\ref{fig:bof_expert}, experts show varying degrees of buffer overfitting, with shallower experts doing slightly better. The average BOF value of the last expert $E_4$ is significantly improved with the help of RSD. This is benefited from the various feature representations learned by all experts. This is consistent with the discussion of the ablation study in Sec.~\ref{sec:main_results}). In other words, knowledge of feature representation in shallower experts $E_{i\leq3}$ is transferred to the final predictor $E_4=F$ and it helps alleviate the buffer overfitting problem of old tasks.
\end{enumerate}

The above statement holds for OCL with different memory sizes, demonstrating the excellent compatibility of multi-level feature learning and self-distillation and showcasing a synergy within MOSE where the sum is greater than its parts.

\end{document}